\documentclass[10pt,twocolumn,letterpaper]{article}

\usepackage[pagenumbers]{cvpr}
\usepackage{amsfonts}
\usepackage{graphicx}
\usepackage{pifont}
\usepackage{tabularx}
\usepackage{multicol}
\usepackage{multirow}
\usepackage{threeparttable}
\usepackage{caption}
\usepackage{xcolor}
\usepackage{makecell}
\usepackage{tcolorbox}
\usepackage{booktabs}
\usepackage{bm}
\usepackage{adjustbox}
\usepackage{placeins}
\usepackage{float}
\usepackage{ragged2e}
\usepackage{etoolbox}
\usepackage{microtype}

\graphicspath{{./}}

\floatstyle{ruled}
\newfloat{algorithm}{tbp}{loa}
\floatname{algorithm}{Algorithm}

\newcounter{selectalgorithmline}
\newenvironment{algorithmic}[1][]{%
  \setcounter{selectalgorithmline}{0}%
  \begin{list}{}{%
    \setlength{\leftmargin}{2.4em}%
    \setlength{\labelwidth}{1.8em}%
    \setlength{\labelsep}{0.4em}%
    \setlength{\itemsep}{0.15em}%
    \setlength{\parsep}{0pt}%
    \setlength{\topsep}{0.25em}%
  }%
}{\end{list}}

\newcommand{\State}{%
  \stepcounter{selectalgorithmline}%
  \item[\scriptsize\arabic{selectalgorithmline}:]%
}
\newcommand{\Input}{\item[\textbf{Input:}]}
\newcommand{\Output}{\item[\textbf{Output:}]}

\newif\ifselectmainmanuscript
\selectmainmanuscriptfalse
\BeforeBeginEnvironment{equation}{%
  \begingroup
  \small
  \let\small\relax
  \let\footnotesize\relax
  \let\scriptsize\relax
  \let\tiny\relax
}
\AfterEndEnvironment{equation}{\endgroup}

\newcommand{\selecttablefont}{\fontsize{7.5pt}{9pt}\selectfont}
\newcommand{\selectmaxwidthbox}[3]{\adjustbox{max width=#1}{#3}}
\newcommand{\selecttablelayout}{%
  \let\sffamily\rmfamily
  \rmfamily
  \let\small\selecttablefont
  \let\footnotesize\selecttablefont
  \let\scriptsize\selecttablefont
  \let\tiny\selecttablefont
  \let\resizebox\selectmaxwidthbox
  \sloppy
  \setlength{\emergencystretch}{1em}%
  \selecttablefont
}
\newcommand{\selectfigurelayout}{%
  \let\sffamily\rmfamily
  \rmfamily
}
\AtBeginEnvironment{table}{\selecttablelayout}
\AtBeginEnvironment{table*}{\selecttablelayout}
\AtBeginEnvironment{figure}{\selectfigurelayout}
\AtBeginEnvironment{figure*}{\selectfigurelayout}
\AtBeginEnvironment{algorithm}{\selectfigurelayout\footnotesize}


\newif\ifselectsupplement
\selectsupplementfalse
\newcounter{selectsupplementitemize}
\begingroup
  \catcode`\/=\active
  \gdef\selectactivatebreakableslash{%
    \catcode`\/=\active
    \def/{\slash}%
  }
\endgroup
\AtBeginEnvironment{itemize}{%
  \ifselectsupplement
    \stepcounter{selectsupplementitemize}%
    \ifnum\value{selectsupplementitemize}=6\relax
      \selectactivatebreakableslash\RaggedRight
    \fi
  \fi
}

\selectmainmanuscripttrue

\definecolor{cvprblue}{rgb}{0.21,0.49,0.74}
\usepackage[breaklinks,colorlinks,allcolors=cvprblue]{hyperref}

\title{SELECT: Semantic Erasure Boundary Construction with Sibling Exclusive Counterfactuals}

\author{Tong Zhang$^{1}$,\;
  Ru Zhang$^{1,*}$,\;
  Jianyi Liu$^{1}$,\;
  Zhen Yang$^{1}$\\
  \normalsize $^{1}$ Beijing University of Posts and Telecommunications, School of Cyberspace Security, Beijing, 100876, China\\
  {\tt\small zenozt@bupt.edu.cn, zhangru@bupt.edu.cn, liujy@bupt.edu.cn, yangzhenyz@bupt.edu.cn}\\
  {\tt\small *Corresponding author}}

\begin{document}

\hypersetup{
  pdftitle={SELECT: Semantic Erasure Boundary Construction with Sibling Exclusive Counterfactuals},
  pdfauthor={Tong Zhang, Ru Zhang, Jianyi Liu, Zhen Yang}
}

\maketitle

\begin{abstract}
Concept erasure aims to suppress undesirable concepts in text-to-image diffusion models while preserving their generative capabilities. Existing methods often redirect target concepts toward predefined semantic anchors, but the influence of anchor selection on the scope of concept erasure remains insufficiently understood. Through causal tracing, we reveal that concepts with more dispersed causal effects tend to exhibit higher anchor sensitivity and lower erasure robustness. Controlled anchor interventions demonstrate that anchor selection affects target suppression and collateral damage to surrounding semantic concepts. These findings motivate us to formulate anchor selection as semantic erasure boundary construction. We propose SELECT (Selective Erasure via Local Exclusive Counterfactual Targets), a plug-and-play framework based on Sibling Exclusive Concepts (SECs), which are semantically close to the target but mutually exclusive in its defining attributes. SELECT jointly constructs a counterfactual redirection endpoint and a local retention set to balance target suppression and semantic preservation. SELECT filters candidates using contextual activation, and uses semantic coherence to identify context compatible redirection endpoints, while the remaining SECs serve as local preservation constraints. Experiments on object, celebrity, artist-style, and NSFW erasure show that SELECT strengthens target suppression over fixed general anchors and achieves suppression-preservation trade-offs, while reducing concept re-emergence and erosion.
\end{abstract}


\section{Introduction}
\label{sec: introduction}
Text-to-image (T2I) diffusion models have demonstrated remarkable capabilities in generating high fidelity and semantically rich images \cite{rombach2022ldm,saharia2022imagen}. However, these models can produce unsafe, copyrighted, private, or otherwise undesirable content learned from large scale training corpora \cite{carlini2023extracting,schramowski2023safe}. Concept erasure has therefore emerged as an important safety editing technique, aiming to remove a target concept from a pre-trained diffusion model while preserving its general generative capability \cite{gandikota2023erasing,kumari2023ablating,heng2023selective,bui2024erasing,wu2025erasing,lee2025localized,wu2025unlearning}.

A broad family of erasure methods can be viewed as a target-to-anchor redirection process \cite{kumari2023ablating,gandikota2024unified,gong2024reliable}. Given a target concept $c_t$ to be erased, the edited model is encouraged to behave as if $c_t$ were replaced by an anchor concept $a$. We refer to $a$ as the anchor, i.e., the semantic endpoint used to redirect the generation behavior originally associated with the target. 

Most existing studies implicitly treat the anchor as a substitute endpoint \cite{zhao2025advanchor,fuchi2024erasing,bui2025fantastic}. Under this view, a good anchor is expected to suppress the target concept by providing an alternative semantic destination. However, this endpoint centric view is incomplete. The anchor is not only a semantic endpoint; it also defines the reference response used by the erasure objective. Changing the anchor therefore changes the model update induced by the same erasure algorithm. Because this update acts on shared model parameters, anchor selection may have consequences beyond target redirection. This raises a fundamental question: does the anchor only determine how the target is suppressed, or does it also shape the collateral effect of erasure on semantically related concepts?

To answer this question, we conduct a systematic diagnostic study of target-to-anchor concept erasure. First, we perform causal tracing over multiple concept categories and observe substantially different causal distribution patterns. Concepts with more dispersed causal effects exhibit higher anchor sensitivity and lower erasure robustness, explaining why fixed anchors cannot reliably fit all concepts. Second, we conduct controlled anchor type interventions. The results show that anchors systematically affect both target suppression and the preservation of semantic neighbors, and that the scope and severity of non-target damage vary with the selected anchor.  

These findings suggest that anchor selection is not a peripheral implementation detail. Instead, the anchor implicitly shapes a semantic erasure boundary. Fixed anchors have inherent limitations because they impose the same redirection endpoint on concepts with different causal structures and semantic neighborhoods. Such fixed mappings cannot provide precise erasure boundaries for heterogeneous concepts, leading to concept re-emergence or erosion. These observations motivate us to rethink anchor design as semantic erasure boundary construction, consisting of a redirection endpoint and a boundary retention set. The endpoint determines where to redirect, whereas the boundary retention set specifies where erasure should stop.

We identify a favorable candidate region for constructing the redirection endpoint. Anchors in this region should be sufficiently close to the target to avoid excessive semantic displacement, while remaining sufficiently exclusive to prevent the retention of target defining attributes. We refer to concepts satisfying this local yet exclusive relationship as Sibling Exclusive Concepts (SECs). Building on SECs, we propose SELECT (Selective Erasure via Local Exclusive Counterfactual Targets), a plug-and-play semantic boundary construction framework. SELECT first uses a large language model to generate SEC candidates, evaluates their overlap with target associated contexts using a contextual activation criterion, and then uses a semantic coherence criterion to prioritize endpoints that are compatible with the original prompt context. We further introduce Anchor-Guided Retain, which organizes the remaining high quality SECs into a retain list and passes them to downstream erasure methods as explicit local preservation constraints.

SELECT automatically constructs a more precise erasure boundary for each target concept, mitigating concept re-emergence and erosion. The entire process performs anchor selection before downstream erasure using lightweight quantitative criteria and does not require downstream editing for candidate evaluation. Independent quantitative evaluation reduces reliance on unvalidated LLM semantic judgments during candidate generation. Our main contributions are summarized as follows:

\begin{itemize}
	\item We experimentally reveal that the anchor implicitly shapes the semantic erasure boundary between target suppression and the preservation of semantic neighbors, and formulate anchor design as semantic erasure boundary construction.
	\item We introduce Sibling Exclusive Concepts and propose SELECT, a plug-and-play semantic boundary construction framework that jointly selects a counterfactual anchor and constructs a boundary retention set.
	\item We validate SELECT across multiple concept categories and erasure methods. The results show that SELECT consistently strengthens target suppression over fixed general anchors and achieves favorable suppression-preservation trade-offs across different erasure settings.
\end{itemize}
\section{Related Work}
\label{sec: related work}

\textbf{Concept erasure}. Research in this area primarily focuses on two paradigms: fine-tuning and closed-form solutions. Fine-tuning methods modify model weights through iterative training or lightweight adapters to suppress the generation of a target concept \cite{lyu2024one,gandikota2023erasing}. These methods perform concept editing through techniques such as training learnable vectors \cite{lyu2024one}, training lightweight erasure modules \cite{lu2024mace,huang2024receler,chen2025growth}, improving classifier-free guidance \cite{hong2024all,kumari2023ablating}, Attention localization\cite{gao2025eraseanything,meng2025concept}, adversarial training \cite{huang2024receler,kim2024race,meng2026dark}, knowledge distillation \cite{kim2024safeguard,xiong2024editing}, multimodal collaboration \cite{li2025one} and continual learning \cite{heng2023selective}. In contrast, closed-form methods erase concepts through direct weight updates without iterative training. SPEED \cite{li2026speed} preserves non-target concepts through Influence-based Prior Filtering and expands prior coverage via Directed Prior Augmentation (DPA). OCE \cite{pmlr-v306-sun26ax} applies closed-form orthogonal transformations for multiplicative weight editing, aiming to suppress targets while preserving parameter geometry. Across these mechanisms, a broad subset of methods can be viewed as redirecting targets toward predefined semantic references, making anchor selection important for erasure behavior.

\textbf{Anchor selection}. Early methods typically use fixed anchors \cite{gandikota2023erasing,kumari2023ablating}. Subsequent studies explore dynamic anchor selection by perturbing embedding vectors \cite{zhao2025advanchor,gong2024reliable} or mapping the target to neighboring concepts \cite{fuchi2024erasing}. More recently, large language models (LLMs) have been used to automatically discover semantically related anchors \cite{xue2025crce}. However, LLM based anchor scoring may inherit semantic biases, while independent evaluation of generated anchors remains limited. The work most closely related to ours is AGE \cite{bui2025fantastic}, which models the concept space as a graph, reveals the locality of erasure effects, and jointly optimizes adaptive target representations and model parameters within a standalone erasure framework. In contrast, SELECT focuses on constructing semantic erasure boundaries before downstream editing. Rather than solely optimizing the redirection target, SELECT jointly determines a counterfactual endpoint and an explicit local retention set, which can be incorporated into different compatible erasure algorithms. Recent work also addresses semantic locality and preservation. NLCE protects neighboring concepts through localized erasure \cite{shi2026neighbor}, ETC anchors target distribution boundaries without predefined anchors \cite{seo2026erasing}, and GrOCE uses dynamic concept graphs and adaptive target clusters \cite{han2026groce}. Closed-Form Concept Erasure via Double Projections \cite{zhang2026closed} uses constrained projections to improve non-target preservation. Different from these approaches, SELECT focuses on how anchor selection itself shapes the semantic scope of target-to-anchor erasure. It jointly constructs a counterfactual redirection endpoint and a local boundary retention set from Sibling Exclusive Concepts, while independently evaluating candidate anchors through contextual activation and semantic coherence.
\begin{figure*}[!htbp]
	\centering 
	\includegraphics[width=1\textwidth]{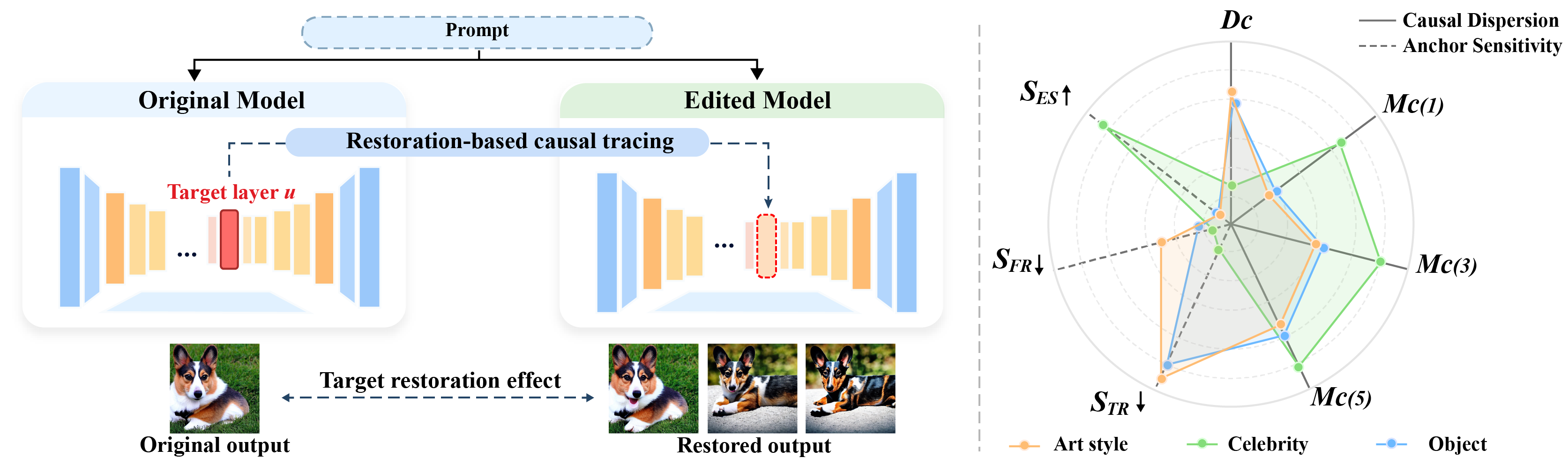} 
	\caption{Causal tracing across concept categories. (a) Restoration-based intervention framework. We inject one cross-attention layer from the erased checkpoint into the original model at a time and measure the resulting change in target response to quantify its causal contribution. (b) Category averaged causal distribution metrics ($D_c$, $M_c(1)$, $M_c(3)$, and $M_c(5)$) and anchor sensitivity metrics ($S_{\mathrm{TR}}$, $S_{\mathrm{FR}}$, and $S_{\mathrm{ES}}$) for artist-style, celebrity, and object concepts.}  
	\label{figure1} 
\end{figure*}
\section{Preliminary}

Concept erasure in text-to-image (T2I) diffusion models aims to remove a target concept from a pre-trained model, preventing the edited model from generating images containing that concept while preserving its general generative capability. Concept erasure methods can be abstracted as a target-to-anchor redirection process \cite{kumari2023ablating,gandikota2024unified,gong2024reliable}: the target concept $c_t$ is redirected toward an anchor concept $a$, such that prompts containing $c_t$ induce model behaviors closer to those associated with $a$. 

Formally, let $M_W$ and $M_{W+\Delta}$ denote the original and edited models, respectively. We abstract target-anchor-based erasure as:
\begin{equation}
	\scriptsize
	\begin{aligned}
	\Delta^\star
	&= \arg\min_{\Delta}\;\Bigl[
	D\big(f_{W+\Delta}(c_t), f_W(a)\big) \\
	&\qquad + \lambda D\big(f_{W+\Delta}(C_r), f_W(C_r)\big)
	\Bigr]
	\end{aligned}
\end{equation}
where $C_r$ denotes the set of other concepts, $f_W(\cdot)$ represents the model behavior, and $D(\cdot,\cdot)$ is a discrepancy measure. We evaluate concept erasure in terms of two basic aspects: target suppression and concept preservation. An ideal erasure method should therefore suppress the target concept as much as possible while preserving the model's behavior on non-target concepts.

\section{Rethinking Anchors as Semantic Boundaries}

In target-to-anchor erasure, the anchor is typically regarded as a substitute semantic endpoint. We argue that this endpoint centric view is incomplete and hypothesize that the anchor implicitly influences the formation of the erasure boundary. We therefore revisit the role of the anchor from two perspectives. First, we use causal tracing to examine whether the internal causal distributions of different concepts are associated with their anchor sensitivity. We then conduct controlled anchor interventions to directly investigate how different semantic types of anchors affect target suppression and local collateral damage.

\subsection{Causal Dispersion and Anchor Sensitivity}
We first examine whether different concepts exhibit different internal erasure patterns. We apply restoration based intervention \cite{meng2022rome} to perform causal tracing over cross-attention units in the diffusion model (Fig.~\ref{figure1}). We prepare the original model $M_{ori}$ and the erased model $M_{edit}$. At each intervention, we replace only one cross-attention unit $u$ in $M_{ori}$ with its corresponding component from $M_{edit}$, obtaining the intervention model $M_u$. For a concept $c$, we measure its target concept response $R_c$ and normalize the positive response reduction associated with each intervention into a causal effect distribution:
\begin{equation}
	\small
	p_c(u)=  
	\frac{  
		\max\left(R_c(M_{ori})-R_c(M_{u}),0\right)  
	}  
	{  
		\sum_{u \in U} \max\left(R_c(M_{ori})-R_c(M_{u}),0\right)  
	} 
\end{equation}
Based on the Shannon entropy of $p_c(u)$ \cite{shannon1948information}, we further compute the normalized causal dispersion metric $D_c$ to characterize how the causal effects of a concept are distributed across units. A smaller $D_c$ indicates that the erasure effect is concentrated in fewer key units:
\begin{equation}
	\small
	D_c=  
	\frac{  
		\exp\left(-\sum_{u \in U} p_c(u)\log p_c(u)\right)  
	}  
	{|\mathcal{U}|}
\end{equation}

We also report the Top $k$ causal mass $M_c(k)$, which measures the proportion of the total erasure related causal contribution covered by the $k$ units with the highest contributions.

To measure anchor sensitivity $S_c$, we evaluate each concept under seven groups of anchors using target residual $S_{\mathrm{TR}}$, erasure strength $S_{\mathrm{ES}}$, and anchor failure rate $S_{\mathrm{FR}}$. A higher average target residual, weaker erasure strength, and higher anchor failure rate indicate greater vulnerability to anchor selection, whereas the opposite indicates more robust erasure performance. We evaluate 100 concepts from each of three categories: artist style, celebrity, and object. Fig.~\ref{figure1}(b) summarizes the causal distribution and each anchor sensitivity metrics. The complete statistics are provided in Appendix B.1.

We observe distinct causal distribution patterns and corresponding differences in erasure performance across concept categories. Artist style and object concepts exhibit relatively dispersed causal effects. In contrast, celebrity concepts show substantially more concentrated causal effects, with their erasure effects dominated by fewer key cross-attention units. Correspondingly, celebrity concepts exhibit lower target residuals and anchor failure rates, together with higher erasure strength, whereas artist-style and object concepts are more susceptible to residual target information under the evaluated anchors. These category differences suggest an association between causal dispersion and anchor dependent erasure behavior.

Importantly, this relationship is not limited to differences between concept categories. To examine variations within each category, we compute the Pearson correlation coefficient $r$ and Spearman correlation coefficient $\rho$ between causal dispersion $D_c$ and anchor sensitivity metric across individual concepts within each category. The results reveal consistent directional associations: higher causal dispersion is positively correlated with target residual and anchor failure rate, but negatively correlated with erasure strength. These findings indicate that concepts with more dispersed causal effects tend to be more susceptible to anchor selection, even within the same semantic category. This comparison is consistent with the directional association identified by the correlation analysis within each category.

\begin{figure}[!htbp]
	\centering 
	\includegraphics[width=1\columnwidth]{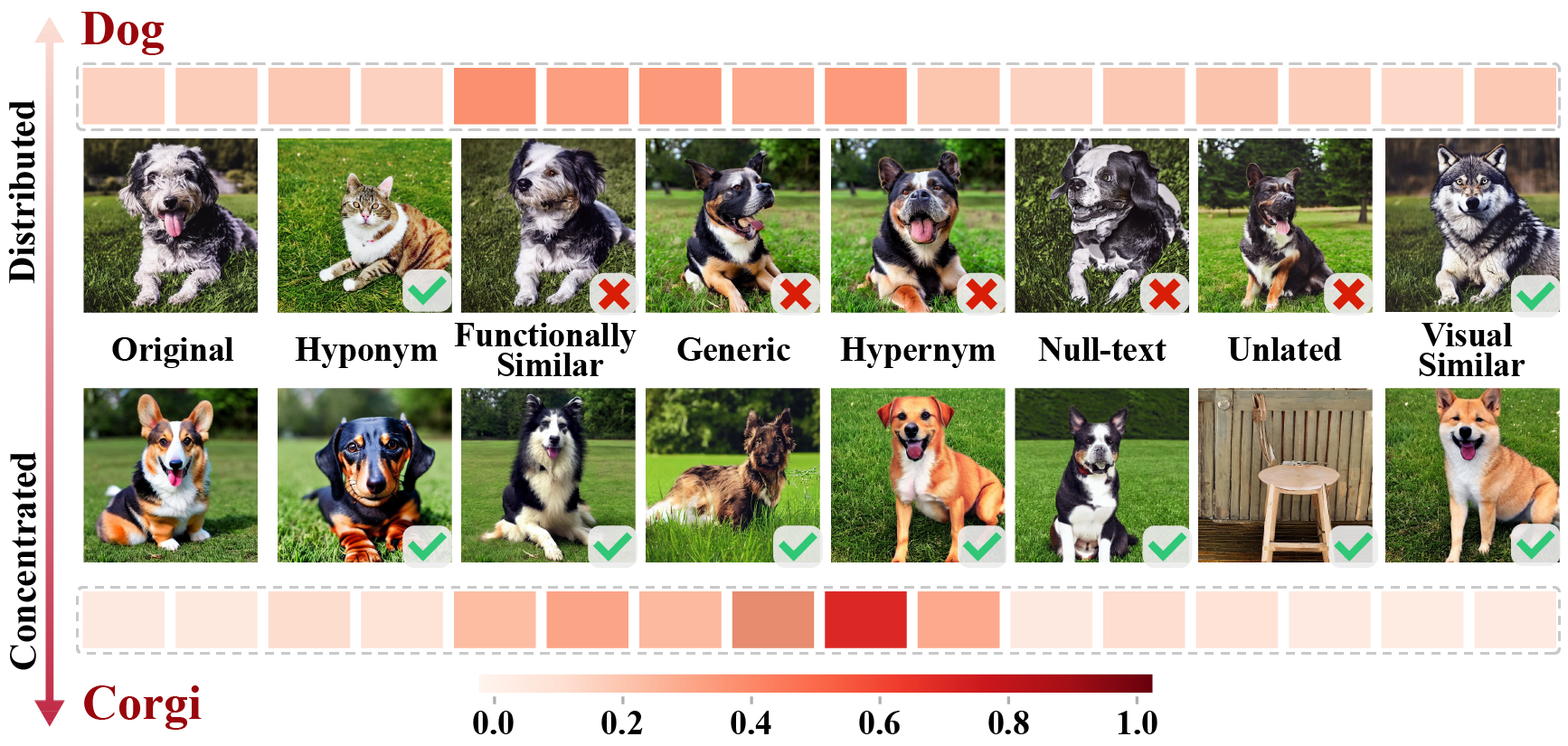} 
	\caption{Consistent relationship between causal dispersion and anchor sensitivity within the object category. Dog shows more dispersed causal effects and greater sensitivity to anchor selection, resulting in less stable target suppression across different anchors. In contrast, Corgi exhibits more concentrated causal effects and more consistent erasure performance. The causal heatmaps and representative generations illustrate that the association between causal dispersion and anchor sensitivity also holds within the same semantic category.}  
	\label{figure2} 
\end{figure}

We further illustrate this association within the object category using Dog and Corgi, as shown in Fig.~\ref{figure2}. General concepts with broader semantic scopes and multiple subordinate visual forms tend to exhibit more dispersed causal effects and higher anchor sensitivity, while more specific concepts with relatively constrained visual characteristics tend to show greater causal concentration and lower anchor sensitivity. 

Overall, the results suggest that anchor sensitivity is associated with the internal causal organization of individual concepts. Consequently, a single fixed anchor may not reliably accommodate concepts with different causal distribution patterns.

\subsection{Controlled Anchor Intervention}
\begin{figure*}[!htbp]
	\centering 
	\includegraphics[width=1\textwidth]{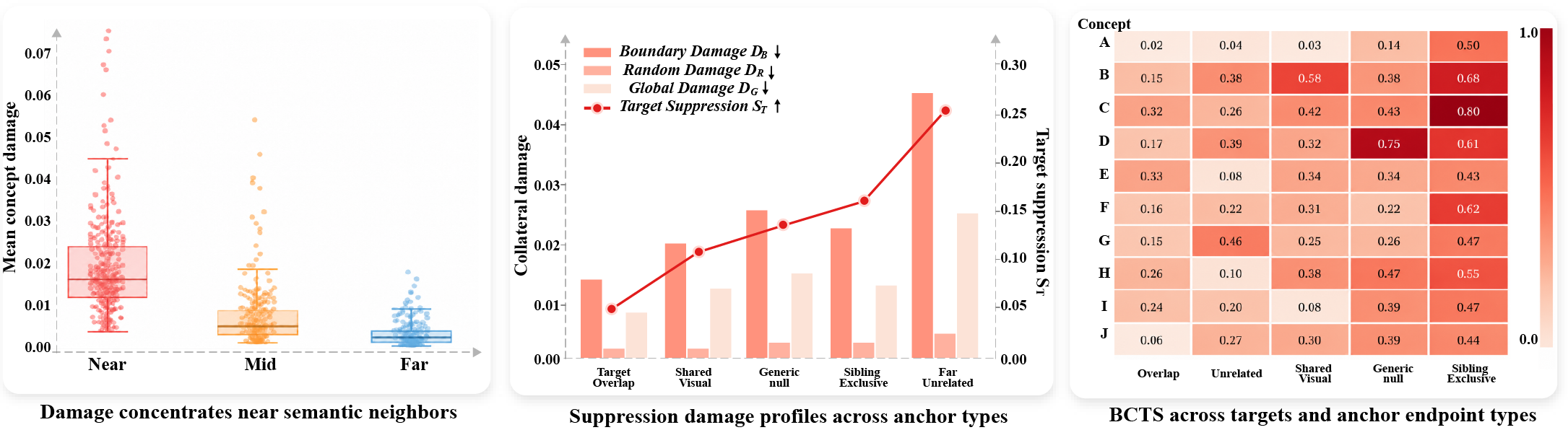} 
	\caption{Anchor selection jointly affects target suppression and the locality of collateral damage. (a) Damage is concentrated around the local semantic neighborhood, with boundary concepts affected more strongly than distant unrelated concepts. (b) Visualization of target suppression and non-target damage under different anchor types. (c) BCTS across ten targets and five anchor types. Sibling Exclusive anchors achieve the strongest overall suppression-preservation trade-off.}  
	\label{figure3} 
\end{figure*}

To further investigate whether anchor selection changes the influence of target erasure on the local semantic space, we conduct controlled anchor interventions on 10 concepts. We fix the erasure algorithm, parameter settings, and all other experimental conditions, changing only the anchor type. We then evaluate both target suppression and the damage to non-target concepts at different semantic ranges. The results are shown in Fig.~\ref{figure3}(a). Results show that boundary concepts suffer substantially greater damage than random unrelated concepts. The complete results are provided in Appendix B.2.

We next examine whether this local spread changes with anchor selection. For each target concept, we consider five anchor types: Target overlap, Shared visual, Generic/null, Far unrelated, and Sibling exclusive. Under otherwise identical experimental settings, we evaluate the erasure performance and the resulting changes in the scope of non-target damage. We additionally measure Global Damage over a broader open concept pool, as reported in Fig.~\ref{figure3}(b). 

The results show that Target overlap anchors provide the weakest target suppression while causing less collateral damage. Far unrelated anchors achieve stronger target suppression but cause the largest boundary damage and global non-target damage. When an anchor is too close to the target, target related attributes are more likely to be retained, resulting in insufficient erasure. In contrast, when the anchor is too far from the target, stronger target suppression is accompanied by greater propagation of the erasure effect into both the local and broader semantic neighborhoods. Anchor choice therefore determines not only the degree of target suppression, but also the scope and severity with which the erasure effect propagates to the non-target semantic neighborhood.

To jointly characterize this suppression preservation trade-off, we further introduce the Boundary Controlled Trade-off Score (BCTS):
\begin{equation}
	\small
	\mathrm{BCTS}(c,a)=\phi_c\left(S(c,a)\right)\left(1-\widetilde{D}(c,a)\right)
\end{equation}
\begin{equation}
	\small
	\phi_c(S)=\min\left(\frac{S}{\tau_c},1\right)
\end{equation}
where $S(c,a)\in[0,1]$ denotes the target suppression strength, $\tau_c$ is the erasure threshold for target concept $c$, and $\widetilde{D}(c,a)$ denotes the combined semantic damage to non-target concepts. BCTS first requires an anchor to achieve sufficient target suppression. Once suppression exceeds the threshold, the score no longer rewards more aggressive erasure and instead primarily evaluates the preservation of non-target semantics. A higher BCTS therefore indicates a better balance between sufficient target suppression and semantic preservation. Fig.~\ref{figure3}(c) reports the BCTS results for five anchor types across ten concepts. Sibling exclusive anchors achieve the highest average BCTS, indicating that the local but exclusive semantic region provides a more stable suppression preservation trade-off overall.

Taken together, these interventions show that the anchor is more than a substitute endpoint. Its semantic relationship with the target jointly affects the suppression outcome and the collateral effects on the surrounding semantic neighborhood, thereby contributing to the functional semantic boundary of concept erasure.

\section{SELECT}

The diagnostic experiments show that the anchor implicitly shapes the erasure boundary between target suppression and neighbor preservation. Anchor selection should therefore not be treated as a secondary implementation detail in concept erasure, but should instead be explicitly designed together with the erasure algorithm and editing strength. However, a counterfactual endpoint primarily specifies where the target should be redirected and does not explicitly determine which non-target semantics around the target should remain unchanged during editing. Because collateral damage is concentrated mainly within the local semantic neighborhood, we further associate the selected anchor with a local retention set that explicitly protects neighboring semantics. We therefore formulate SELECT as a framework for semantic boundary construction. For a target concept $c_t$, SELECT constructs a semantic boundary specification:
\begin{equation}
	\Gamma_t=(a_t^\star,B_t)
\end{equation}
where $a_t^\star$ is the counterfactual anchor that redirects the target concept, and $B_t$ is the boundary retention set that constrains the preservation of vulnerable neighboring concepts.

\subsection{SEC Candidate Construction}
To jointly support target suppression and neighbor preservation, we require an anchor to satisfy two constraints: local semantic proximity and core attribute exclusivity. Local semantic proximity requires the anchor to lie within the local semantic neighborhood of the target. When the anchor is too far from the target, the resulting redirection introduces a large semantic shift and causes greater damage to concepts in the broader semantic space. Core attribute exclusivity requires the anchor not to share target specific attributes, thereby reducing the risk of retaining the defining attributes of the target. 

We define Sibling Exclusive Concepts (SECs) as the candidate anchor region for SELECT. An SEC shares the same parent semantic category as the target, providing a smooth redirection path, while remaining mutually exclusive with the target in its core attributes, thereby providing a clear counterfactual endpoint. A schematic illustration of the semantic regions used for anchor construction is provided in Appendix A. Given a concept space $C$ and its semantic hierarchy $H$, the SEC candidate set for a target concept $c_t$ is defined as:
\begin{equation}
	\scriptsize
	\begin{aligned}
	SEC(c_t)=\{a\in C \mid {}&parent_H(a)=parent_H(c_t),\ a\neq c_t, \\
	&\ Excl(a,c_t)=1\}.
	\end{aligned}
\end{equation}
Here, $parent_H(\cdot)$ denotes the parent node of a concept in the semantic hierarchy $H$, and $Excl(a,c_t)=1$ indicates that anchor $a$ and target $c_t$ are mutually exclusive in their core attributes.

\begin{figure}[t] 
	\centering 
	\includegraphics[width=1\columnwidth]{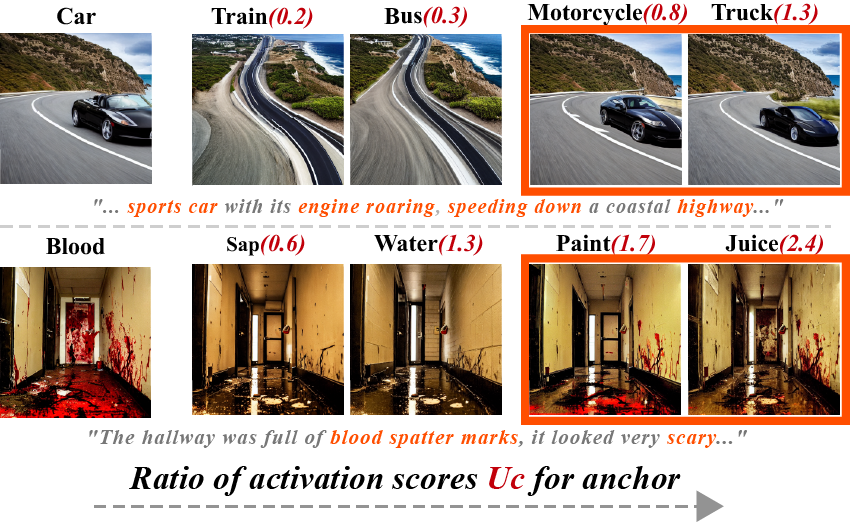} 
	\caption{Effect of contextual activation on SEC anchor quality. For representative Car and Blood targets, low $U(a,c_t)$ anchors suppress target associated cues under related context prompts, whereas high $U(a,c_t)$ anchors retain stronger target related visual evidence and are more prone to re-emergence.}  
	\label{figure4} 
\end{figure}

\subsection{Anchor Evaluation}
Although SECs define a favorable anchor region, SEC candidates generated manually or by an LLM still require quantitative evaluation. We therefore introduce two criteria for anchor evaluation: contextual activation and semantic coherence.

\subsubsection{Contextual Activation}
Prior studies show that erased concepts can re-emerge under alternative prompts, adversarial conditions, or residual latent representations \cite{pham2024circumventing,rusanovsky2025memories,srivatsan2025stereo}. We further examine whether ordinary target associated contexts can induce a similar reactivation effect. Even after the target is redirected to another anchor, context words that are strongly associated with the target may reactivate residual target representations within the model. For example, after erasing \{blood\}, a prompt containing strongly associated context words such as \{red, wound, stain\} may still generate blood stains or red stain like patterns. This suggests that concept erasure may overlook the association between a concept and its surrounding semantic context.

\begin{figure}[t] 
	\centering 
	\includegraphics[width=1\columnwidth]{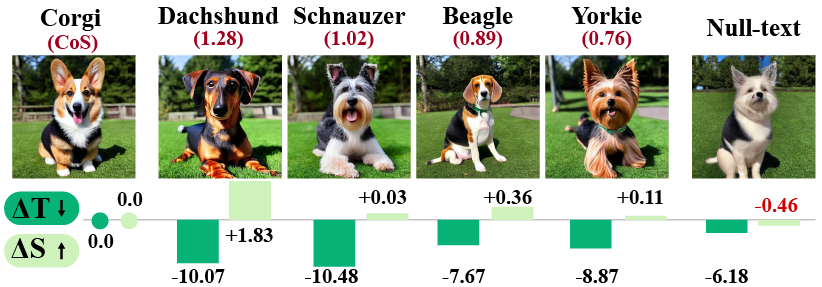} 
	\caption{Semantic coherence and the suppression preservation trade-off. Candidate anchors for Corgi erasure are compared using $CoS$, target CLIP change $\Delta T$, and scene CLIP change $\Delta S$. Higher coherence anchors provide stronger target suppression while preserving more of the non-target scene content.}  
	\label{figure5} 
\end{figure}

We first examine whether target associated contexts indeed increase activation of the target concept. For a target concept $c_t$, we construct related context templates $T_{rel}$ and neutral context templates $T_{neu}$. The former contain contextual cues strongly associated with $c_t$, whereas the latter describe neutral scenes. We use a masked language model \cite{devlin2019bert} to estimate the probability of predicting $c_t$ at the $MASK$ position under different templates:
\begin{equation}
	\scriptsize
	\begin{aligned}
	A_{\tau}(c_t)
	&=\frac{1}{|T_{\tau}|}
	\sum_{T\in T_{\tau}}
	P_{MLM}(c_t\mid T([MASK])) \\
	&\qquad \tau\in\{\mathrm{rel},\mathrm{neu}\}
	\end{aligned}
\end{equation}
We then define the Context Raise Ratio $CRR(c_t)$:
\begin{equation}
	\small
	CRR(c_t)=  
	\frac{A_{rel}(c_t)}  
	{A_{neu}(c_t)+\epsilon}  
\end{equation}
$CRR(c_t)$ measures the additional activation induced by related contexts relative to neutral contexts. We further generate corresponding images for validation. We use CLIP Top-1 classification \cite{radford2021clip} to compute the concept re-emergence rate $R(c_t)$, defined as the proportion of generated images in which the erased target concept reappears. We found that, concepts with more frequent re-emergence exhibit substantially stronger activation under related context prompts than under neutral contexts. The complete results are provided in Appendix B.

\begin{figure*}[!htbp]
	\centering 
	\includegraphics[width=1\textwidth]{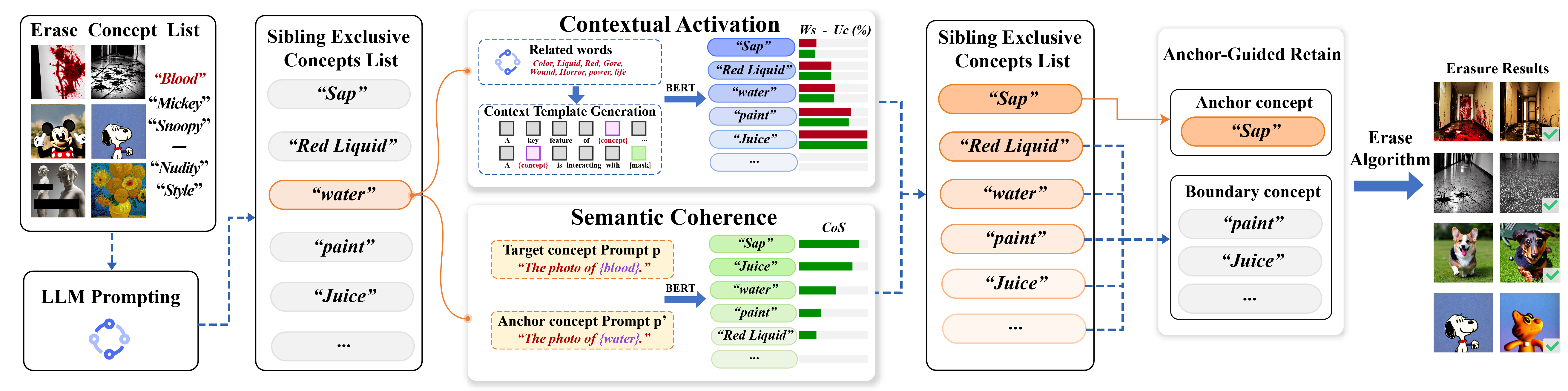} 
	\caption{Overview of SELECT. SELECT first generates Sibling Exclusive Concept candidates using an LLM, filters them using contextual activation, and ranks the remaining candidates by semantic coherence. The selected counterfactual anchor determines the redirection endpoint, while the remaining high quality SECs form the boundary retention set passed to the downstream erasure algorithm.}  
	\label{figure6} 
\end{figure*}

Based on this observation, we use contextual activation to evaluate SEC anchors. For a target $c_t$, we use an LLM with templates such as  \{When people think of {$c_t$}, they think of [MASK]\} to generate associated words and construct the target associated word set $W_t$. We then compute the activation probability for $x\in \{c_t,a\}$ with respect to each associated word $w_i\in W_t$:
\begin{equation}
	\small
	W(x,w_i)=  
	\frac{1}{|T_{rel}|}  
	\sum_{T\in T_{rel}}  
	P_{MLM}(w_i\mid T(x,[MASK]))
\end{equation}
To reduce noise from weakly related words, we retain only the Top-$K$ associated words with the highest activation scores for target $c_t$:
\begin{equation}
	\small
	W_t^{K}=  
	TopK_{w\in W_t}\big(W(c_t,w),K\big) 
\end{equation}
For a candidate concept $a$, we compute its mean activation score over the selected associated words:
\begin{equation}
	\small
	S(a,c_t)=  
	\frac{1}{|W_t^{K}|}  
	\sum_{w\in W_t^{K}}  
	W(a,w)
\end{equation}
Finally, we define the contextual activation ratio between candidate anchor $a$ and target $c_t$ as:
\begin{equation}
	\small
	U(a,c_t)=  
	\frac{S(a,c_t)}  
	{S(c_t,c_t)+\epsilon} 
\end{equation}

This ratio characterizes the relative activation of anchor $a$ with respect to target $c_t$ under target-associated contexts, thereby measuring their contextual activation overlap. We examine the activation ratios $U(a,c_t)$ of different SECs together with their generated images in Fig.~\ref{figure4}. Anchors with higher erasure effectiveness tend to have lower $U(a,c_t)$, indicating weaker overlap with the target activation pattern and lower association with target related contextual cues. In contrast, anchors with higher $U(a,c_t)$ are more likely to retain target related cues after editing, increasing the likelihood of target re-emergence.

\subsubsection{Semantic Coherence}
An ideal anchor should not only facilitate concept erasure, but also preserve the visual coherence of the original prompt and avoid visual distortion, background corruption, or loss of non-target content. We attribute these failures to ambiguity and instability in the redirection endpoint. To quantify this linguistic compatibility, we introduce the semantic coherence score $CoS$. Given a target prompt set $P_t$, for each prompt $p\in P_t$, we replace the target concept $c_t$ with a candidate anchor $a$ to obtain:
\begin{equation}
	\small
	p^a=p[c_t\rightarrow a]  
\end{equation}
We define semantic coherence using language model perplexity:
\begin{equation}
	\small
	CoS(a,c_t)=  
	\frac{1}{|P_t|}  
	\sum_{p\in P_t}  
	\frac{PPL(p)}  
	{PPL(p^a)+\epsilon}  
\end{equation}
If the substituted prompt $p^a$ remains natural, fluent, and compatible with the original context, $PPL(p^a)$ is lower and $CoS(a,c_t)$ is higher. A high semantic coherence score therefore indicates that the anchor integrates naturally into the original context and redirects the model toward a stable and well defined semantic region, reducing the likelihood of distorted generations. Fig.~\ref{figure5} illustrates this difference. Anchors with higher semantic coherence provide stronger erasure while better preserving non-target visual elements that are otherwise vulnerable to corruption or reconstruction during concept erasure.

\subsection{Boundary Retention}
Selecting a counterfactual anchor improves the redirection endpoint, but it does not by itself impose an explicit constraint on the neighboring concepts most vulnerable to erasure spillover. Based on this observation, we introduce Anchor-Guided Retain. The core idea is to use SECs to form a critical local semantic boundary. These concepts are close to the target concept but remain distinct from it, making them more likely to be among the concepts most vulnerable to erasure. After selecting the counterfactual anchor for erasure, the highest ranked remaining SECs are added to the retain list of the concept erasure algorithm as boundary concepts:
\begin{equation}
	B_t=TopM(S_{act}\setminus\{a_t^\star\},CoS_t,M)
\end{equation}
Here, $\mathcal{S}_{act}$ denotes the SEC candidates retained after filtering, and $a_t^\star$ is the selected anchor. $\operatorname{TopM}(\cdot)$ ranks the remaining candidates according to $CoS_t$ and selects the top $M$ concepts to form the boundary retention set $B_t$. This boundary concept constraint is designed to mitigate concept erosion by constructing a semantic retention region around the target concept. Through this explicit boundary constraint, the model is guided to erase the target concept more precisely rather than damaging more broadly shared features. We emphasize that preserving these boundary concepts is fundamentally different from using them as redirection targets for erasure. SELECT deliberately avoids the latter to prevent concept re-emergence.

\begin{algorithm}[t]
	\caption{SELECT Semantic Boundary Construction}
	\label{alg1}
	
	\begin{algorithmic}[1]
		
		\Input Target concept $c_t$; LLM generator $G_{LLM}$; 
		SEC prompt template $\mathcal{P}_{SEC}$; context word set $\mathcal{W}_t$; 
		context template set $\mathcal{T}$; target prompt set $\mathcal{P}_t$; 
		retain budget $M$; erasure algorithm $\mathcal{E}$; contextual filtering budget $k$;
		original retain set $C_r$; original model $M$.
		
		\Output counterfactual anchor $a_t^\star$; boundary retention set $B_t$; 
		edited model $M'$.
		
		\State Generate initial SEC candidates:
		$\mathcal{S}_t = G_{LLM}(c_t,\mathcal{P}_{SEC})$
		
		\State Compute the contextual activation:
		$U_t=\{U(a,c_t)\mid a\in\mathcal{S}_t\}$
		
		\State $\mathcal{S}_{act}=TopK_{low}(\mathcal{S}_t,U_t,k)$
		
		\State Compute semantic coherence:
		$CoS_t=\{CoS(a,c_t)\mid a\in\mathcal{S}_{act}\}$
		
		\State Select anchor:
		$a_t^\star=\underset{a\in S_{act}}{\arg\max}\ CoS(a,c_t)$
		
		\State Construct the boundary retention set:
		$B_t=TopM(S_{act}\setminus\{a_t^\star\},CoS_t,M)$
		
		\State Apply the erasure algorithm:
		$M'=\mathcal{E}(M,c_t,a_t^\star,C_r\cup B_t)$
		
		\State Return $\Gamma_t=(a_t^\star,B_t)$ and $M'$.
		
	\end{algorithmic}
\end{algorithm}

\subsection{SELECT}
In summary, SELECT decomposes semantic boundary construction into three steps: 1) generating the SEC candidate set, 2) evaluating anchors, and 3) constructing the retain list. In this process, we use an LLM to automatically generate an SEC candidate set for each target. The complete prompt template is provided in Appendix. The complete SELECT pipeline is illustrated in Algorithm \ref{alg1} and Fig.~\ref{figure6}.

\section{Experiments}
\label{sec: experiments}

% ============================================================
% Header macro for paired object concepts
% ============================================================

\newcommand{\objpairheader}[2]{%
	\hline
	
	& \multicolumn{11}{c|}{\textbf{#1}}
	& \multicolumn{11}{c}{\textbf{#2}} \\
	\cline{2-23}
	
	\multirow{2}{*}{\textbf{Method}}
	
	& \multicolumn{2}{c|}{target}
	& \multicolumn{2}{c|}{synonym}
	& \multicolumn{2}{c|}{random}
	& \multicolumn{2}{c|}{boundary}
	& \multirow{2}{*}{$H_o \uparrow$}
	& \multirow{2}{*}{$H_b \uparrow$}
	& \multirow{2}{*}{BCTS $\uparrow$}
	
	& \multicolumn{2}{c|}{target}
	& \multicolumn{2}{c|}{synonym}
	& \multicolumn{2}{c|}{random}
	& \multicolumn{2}{c|}{boundary}
	& \multirow{2}{*}{$H_o \uparrow$}
	& \multirow{2}{*}{$H_b \uparrow$}
	& \multirow{2}{*}{BCTS $\uparrow$}
	\\
	
	\cline{2-9}
	\cline{13-20}
	
	&
	$\mathrm{Acc}_e \downarrow$ & $cs \downarrow$
	& $\mathrm{Acc}_g \downarrow$ & $cs \downarrow$
	& $\mathrm{Acc}_s \uparrow$ & $cs \uparrow$
	& $\mathrm{Acc}_b \uparrow$ & $cs \uparrow$
	& & &
	&
	$\mathrm{Acc}_e \downarrow$ & $cs \downarrow$
	& $\mathrm{Acc}_g \downarrow$ & $cs \downarrow$
	& $\mathrm{Acc}_s \uparrow$ & $cs \uparrow$
	& $\mathrm{Acc}_b \uparrow$ & $cs \uparrow$
	& & &
	\\
	
	\hline
}

% ============================================================
% Table 3
% ============================================================

\begin{table*}[t]
	\centering
	
	\caption{
		Quantitative evaluation of object erasure under SPEED.
		Nine target concepts and the overall mean are reported.
		Accuracy based metrics, $H_o$, $H_b$, and BCTS are shown on a 0--100 scale.
		Bold values indicate the best result among Null, General, and Ours.
	}
	\label{tab3}
	
	\scriptsize
	\setlength{\tabcolsep}{1.25pt}
	\renewcommand{\arraystretch}{1.12}
	
	\resizebox{\textwidth}{!}{%
		\begin{tabular}{
				l|
				cc|cc|cc|cc|ccc|
				cc|cc|cc|cc|ccc
			}
			
			% ============================================================
			% Cat + Dog
			% ============================================================
			
			\objpairheader{Cat}{Dog}
			
			SD 1.4
			& 100.00 & 27.73
			& 100.00 & 28.36
			& 98.60 & 30.14
			& 99.80 & 30.28
			& -- & -- & --
			& 100.00 & 27.06
			& 64.00 & 27.69
			& 97.00 & 30.03
			& 99.80 & 31.09
			& -- & -- & -- \\
			
			Null
			& 85.00 & 27.35
			& 84.33 & 26.42
			& 96.60 & 29.99
			& 99.40 & 30.35
			& 21.89 & 21.94 & 18.00
			& 97.50 & 26.61
			& 49.00 & 26.01
			& 92.00 & \textbf{29.55}
			& 99.20 & 31.00
			& 7.00 & 7.00 & 2.90 \\
			
			General
			& 92.00 & 27.87
			& 87.33 & 27.53
			& \textbf{97.80} & 30.02
			& \textbf{99.80} & \textbf{30.37}
			& 14.01 & 14.02 & 9.76
			& 74.50 & 26.05
			& 51.67 & 26.02
			& 89.40 & 28.91
			& 96.00 & 30.37
			& 42.20 & 42.70 & 28.10 \\
			
			Ours
			& \textbf{0.00} & \textbf{21.01}
			& \textbf{7.00} & \textbf{22.57}
			& \textbf{97.80} & \textbf{30.07}
			& 99.60 & 30.31
			& \textbf{96.84} & \textbf{97.43} & \textbf{97.40}
			& \textbf{3.00} & \textbf{22.81}
			& \textbf{3.67} & \textbf{24.03}
			& \textbf{94.00} & 29.55
			& \textbf{99.60} & \textbf{31.11}
			& \textbf{95.70} & \textbf{97.60} & \textbf{93.60} \\

			% ============================================================
			% Pig + Corgi
			% ============================================================
			
			\objpairheader{Pig}{Corgi}
			
			SD 1.4
			& 100.00 & 29.21
			& 99.67 & 29.69
			& 93.60 & 29.48
			& 87.20 & 30.30
			& -- & -- & --
			& 100.00 & 33.38
			& 100.00 & 33.27
			& 99.00 & 29.82
			& 95.60 & 31.81
			& -- & -- & -- \\
			
			Null
			& 1.50 & 24.12
			& 32.67 & 24.89
			& 93.00 & 29.51
			& \textbf{87.60} & \textbf{30.27}
			& 84.41 & 82.87 & 80.60
			& 17.50 & 25.81
			& 7.67 & 24.28
			& 99.20 & \textbf{30.03}
			& 90.40 & 31.70
			& 90.82 & 88.20 & 89.60 \\
			
			General
			& 0.50 & 23.99
			& 35.33 & 25.01
			& 90.00 & 29.43
			& \textbf{87.60} & 30.17
			& 83.13 & 82.32 & 77.20
			& 8.00 & 22.68
			& 17.33 & 23.29
			& 99.40 & 29.75
			& 92.00 & 31.63
			& 90.84 & 88.66 & 91.40 \\
			
			Ours
			& \textbf{0.00} & \textbf{22.18}
			& \textbf{0.00} & \textbf{23.42}
			& \textbf{95.60} & \textbf{29.58}
			& 85.60 & 30.26
			& \textbf{98.49} & \textbf{94.69} & \textbf{81.20}
			& \textbf{0.00} & \textbf{22.23}
			& \textbf{0.00} & \textbf{21.39}
			& \textbf{99.80} & 29.59
			& \textbf{92.20} & \textbf{31.76}
			& \textbf{99.93} & \textbf{97.26} & \textbf{92.00} \\

			% ============================================================
			% Garbage truck + Banana
			% ============================================================
			
			\objpairheader{Garbage truck}{Banana}
			
			SD 1.4
			& 91.50 & 29.07
			& 91.33 & 29.81
			& 99.80 & 29.60
			& 88.00 & 29.69
			& -- & -- & --
			& 98.00 & 29.51
			& 99.33 & 30.89
			& 98.40 & 29.08
			& 91.40 & 29.53
			& -- & -- & -- \\
			
			Null
			& 78.50 & 28.37
			& 81.00 & 28.63
			& \textbf{100.00} & 29.53
			& \textbf{89.00} & 29.70
			& 27.49 & 27.18 & 23.92
			& 6.00 & 24.01
			& 16.33 & 25.16
			& 97.60 & 29.01
			& 91.40 & \textbf{29.54}
			& 91.36 & 89.47 & 89.00 \\
			
			General
			& 39.50 & 26.20
			& 63.00 & 28.07
			& 99.80 & \textbf{29.64}
			& 87.60 & \textbf{29.71}
			& 56.00 & 54.57 & 66.10
			& 2.00 & 22.51
			& 12.00 & 24.04
			& 97.60 & 29.04
			& 90.80 & 29.50
			& 94.30 & 92.08 & 88.40 \\
			
			Ours
			& \textbf{0.00} & \textbf{23.85}
			& \textbf{22.00} & \textbf{26.07}
			& \textbf{100.00} & 29.53
			& 87.60 & 29.68
			& \textbf{91.41} & \textbf{87.63} & \textbf{87.60}
			& \textbf{0.00} & \textbf{21.57}
			& \textbf{0.00} & \textbf{22.29}
			& \textbf{98.20} & \textbf{29.14}
			& \textbf{91.60} & 29.53
			& \textbf{99.39} & \textbf{97.03} & \textbf{89.80} \\

			% ============================================================
			% Motorcycle + Watermelon
			% ============================================================
			
			\objpairheader{Motorcycle}{Watermelon}
			
			SD 1.4
			& 96.00 & 28.77
			& 97.00 & 28.89
			& 98.60 & 29.76
			& 89.60 & 29.69
			& -- & -- & --
			& 99.50 & 31.92
			& 99.67 & 31.82
			& 96.00 & 29.83
			& 85.80 & 30.16
			& -- & -- & -- \\
			
			Null
			& 74.50 & 28.65
			& 81.67 & 29.09
			& 95.80 & \textbf{29.73}
			& 87.40 & 29.71
			& 28.79 & 28.52 & 26.92
			& 15.50 & 24.20
			& 44.33 & 25.53
			& 92.20 & 29.72
			& 81.20 & 29.77
			& 76.17 & 73.27 & 72.80 \\
			
			General
			& 62.50 & 28.40
			& 63.33 & 29.01
			& 96.20 & 29.69
			& 89.80 & \textbf{29.72}
			& 46.63 & 46.10 & 40.93
			& 23.00 & 24.86
			& 33.33 & 25.15
			& 92.60 & \textbf{29.84}
			& 83.80 & 29.66
			& 78.11 & 75.76 & 73.62 \\
			
			Ours
			& \textbf{1.50} & \textbf{27.25}
			& \textbf{32.00} & \textbf{28.60}
			& \textbf{97.60} & 29.69
			& \textbf{90.80} & 29.53
			& \textbf{85.46} & \textbf{83.63} & \textbf{88.40}
			& \textbf{0.00} & \textbf{20.84}
			& \textbf{3.33} & \textbf{21.59}
			& \textbf{94.60} & 29.81
			& \textbf{87.40} & \textbf{29.78}
			& \textbf{97.04} & \textbf{94.30} & \textbf{81.80} \\

			% ============================================================
			% Spider-Man + Overall Mean
			% ============================================================
			
			\objpairheader{Spider-Man}{Overall Mean}
			
			SD 1.4
			& 100.00 & 30.16
			& 100.00 & 30.57
			& 94.60 & 29.93
			& 98.60 & 30.70
			& -- & -- & --
			& 98.33 & 29.64
			& 94.56 & 30.11
			& 97.29 & 29.74
			& 92.87 & 30.36
			& -- & -- & -- \\
			
			Null
			& 37.50 & 24.17
			& 48.33 & 24.59
			& 84.60 & 29.07
			& 96.60 & 30.37
			& 63.59 & 65.64 & 63.44
			& 45.94 & 25.92
			& 49.48 & 26.07
			& 94.56 & 29.57
			& 91.36 & 30.27
			& 54.61 & 53.79 & 51.91 \\
			
			General
			& 18.50 & 23.79
			& 33.33 & 24.57
			& 89.00 & 29.34
			& 97.20 & 30.12
			& 77.91 & 79.88 & 86.20
			& 35.61 & 25.15
			& 44.07 & 25.86
			& 94.64 & 29.52
			& 91.62 & 30.14
			& 64.79 & 64.01 & 62.41 \\
			
			Ours
			& \textbf{0.00} & \textbf{21.61}
			& \textbf{0.67} & \textbf{22.61}
			& \textbf{95.20} & \textbf{29.78}
			& \textbf{98.40} & \textbf{30.55}
			& \textbf{98.13} & \textbf{99.24} & \textbf{93.60}
			& \textbf{0.50} & \textbf{22.59}
			& \textbf{7.63} & \textbf{23.62}
			& \textbf{96.98} & \textbf{29.64}
			& \textbf{92.53} & \textbf{30.28}
			& \textbf{95.82} & \textbf{94.31} & \textbf{89.49} \\
			
			\hline
			
		\end{tabular}%
	}
	
\end{table*}

\begin{figure*}[t] 
	\centering 
	\includegraphics[width=1\textwidth]{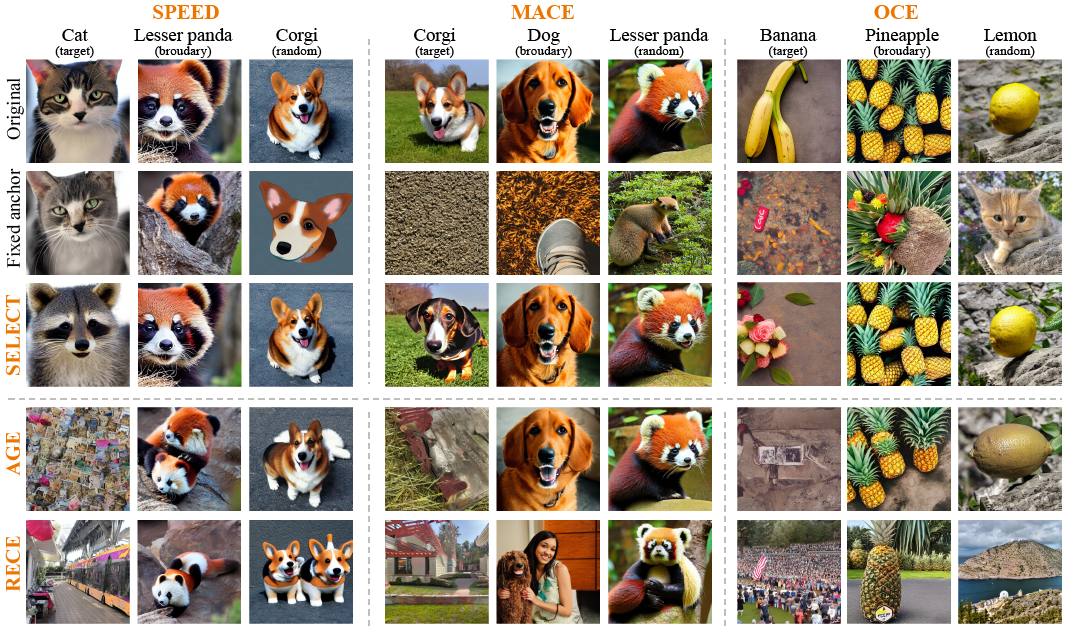} 
	\caption{Qualitative comparison of object erasure under SPEED \cite{li2026speed}, MACE \cite{lu2024mace}, and OCE, together with two standalone erasure frameworks, AGE and RECE. For the three downstream erasure methods, we compare the original model, fixed-anchor editing, and SELECT-enhanced editing. AGE and RECE are evaluated independently on corresponding target and non-target concepts. SELECT effectively suppresses target-specific characteristics while preserving recognizable non-target content in the illustrated examples, whereas alternative methods may exhibit incomplete suppression, semantic drift, or structural degradation.}  
	\label{figure7} 
\end{figure*}

In this section,we evaluate SELECT with three downstream erasure algorithms, SPEED \cite{li2026speed}, MACE \cite{lu2024mace}, and OCE \cite{pmlr-v306-sun26ax}. We further include AGE \cite{bui2025fantastic} and RECE \cite{gong2024reliable} as standalone erasure baselines for qualitative comparison. All experiments are conducted on Stable Diffusion v1.4 \cite{rombach2022ldm}. We use DeepSeek \cite{bi2024deepseek} for SEC candidate generation and auxiliary keyword and template generation. All baseline reproductions and experiments are conducted on an NVIDIA RTX A6000. We evaluate SELECT on four tasks: object, celebrity, artist-style, and NSFW erasure.

\subsection{Object erasure}
In this section, we comprehensively evaluate the performance of SELECT on object erasure in terms of two core dimensions: erasure effectiveness and content retention. We consider a broad range of concepts from different categories. We select 9 concepts spanning animals, vehicles, fruits, and fictional characters for evaluation. For each target, the evaluation covers 15-19 concept conditions, including the target itself, 3-5 boundary concepts, eight random non-target concepts, and 3-5 synonyms or variants. We generate 200 images for each condition. We use two fixed-anchor baselines for comparison: Null-Text and General (e.g., “Sky” and “Ground”).

\textbf{Erasure effectiveness}. $\mathrm{Acc}_e$ is the CLIP top-1 classification accuracy \cite{radford2021clip} on the erased target and $\mathrm{Acc}_g$ is the corresponding accuracy on synonyms/variants. Lower values indicate stronger direct erasure and better generalization.

\textbf{Content retention}. $\mathrm{Acc}_s$ denotes the classification accuracy on the other unrelated concepts among the other test concepts. We additionally use an LLM to generate 3-5 local neighboring concepts for each target and compute $\mathrm{Acc}_b$. Higher $\mathrm{Acc}_s$ and $\mathrm{Acc}_b$ indicate better preservation. We further compute the harmonic mean $H_o$ \cite{lu2024mace} to balance erasure and retention:
$$H_o = \frac{3}{(1 - \text{Acc}_e)^{-1} + (\text{Acc}_s)^{-1} + (1 - \text{Acc}_g)^{-1}}$$
We additionally introduce $H_b$ to evaluate the preservation of local neighboring concepts:
$$H_b = \frac{3}{(1 - \text{Acc}_e)^{-1} + (\text{Acc}_b)^{-1} + (1 - \text{Acc}_g)^{-1}}$$
We also report the Boundary Controlled Trade-off Score (BCTS) introduced in the diagnostic experiments to jointly evaluate target concept erasure and boundary concept preservation.

We first compare SELECT with fixed anchor baselines under SPEED. As shown in Tab.~\ref{tab3}, SELECT provides substantially stronger target and synonym suppression while maintaining high accuracy on random and boundary concepts, resulting in the strongest overall suppression-preservation trade-off.

We further evaluate SELECT with SPEED \cite{li2026speed}, MACE \cite{lu2024mace}, and OCE \cite{pmlr-v306-sun26ax}, which support concept redirection through fixed anchors. By replacing these fixed references with SELECT selected counterfactual anchors and incorporating explicit boundary retention constraints, we examine the effectiveness of SELECT across different editing mechanisms. For comparison, we additionally include AGE \cite{bui2025fantastic} and RECE \cite{gong2024reliable} as standalone erasure frameworks. AGE jointly optimizes adaptive redirection targets and model parameters, whereas RECE identifies and suppresses concept-recovering embeddings through its own editing procedure. Neither requires an external anchor selection stage. Fig.~\ref{figure7} presents the results of the three downstream algorithms alongside the independent AGE and RECE baselines.

Under SPEED \cite{li2026speed}, fixed anchor editing leaves visible target characteristics, whereas SELECT redirects the target toward a distinct semantic concept while preserving non-target generations. Under MACE \cite{lu2024mace}, fixed anchor editing causes substantial structural degradation, while SELECT produces a recognizable non-target dog breed and retains neighboring concepts. For OCE \cite{pmlr-v306-sun26ax}, both fixed-anchor editing and SELECT suppress the banana concept, but SELECT better preserves the semantic identity of non-target fruits. AGE \cite{bui2025fantastic} and RECE \cite{gong2024reliable} also achieve visible target suppression, although some generated images exhibit semantic drift or loss of recognizable object structure. These qualitative results suggest that SELECT can improve the suppression preservation trade-off across different editing mechanisms. 

These results clarify the complementary roles of the two components in SELECT. The selected anchor $a_t^\star$ specifies the counterfactual redirection endpoint, whereas $B_t$ explicitly constrains the local semantic region that should remain unchanged. Together, they form the semantic erasure boundary.

\subsection{Celebrity Erasure}
In this section, we evaluate large-scale celebrity erasure. We selected 200 celebrity concepts, divided into an erased group and a retained group \cite{lu2024mace}. Similar to the previous section, we use SELECT to select a counterfactual anchor for each celebrity concept, adding Null-text and a universal fixed general anchor “a person” for comparison. The generated images are recognized by the GIPHY Celebrity Detector \cite{hasty2019giphy}. We calculate Top1, Top3, and Top5 classification accuracies to test the thoroughness of the erasure and the degree of retention of the celebrities. To jointly measure celebrity erasure efficacy and identity preservation, we compute the harmonic mean $H_c$ using the Top-1 accuracy of the erased celebrities $\mathrm{Acc}_e$ and that of the retained celebrities $\mathrm{Acc}_s$ \cite{lu2024mace}:
\begin{equation}
	\small
	H_c=
	\frac{2}
	{(1-\mathrm{Acc}_e)^{-1}+(\mathrm{Acc}_s)^{-1}}.
\end{equation}

The experimental results are shown in Tab.~\ref{tab4}. SELECT consistently improves erasure effectiveness, achieving lower Top-1, Top-3, and Top-5 detection rates than the corresponding fixed general anchor under both SPEED \cite{li2026speed} and MACE \cite{lu2024mace}. Under SPEED \cite{li2026speed}, SELECT also improves $H_c$ from 81.99 to 87.99, while under MACE \cite{lu2024mace} it achieves stronger erasure with a small reduction in retention accuracy, resulting in $H_c=86.33$ compared with 87.30 for the general anchor.

\subsection{Artist Style Erasure}

\begin{figure}[t] 
	\centering 
	\includegraphics[width=1\columnwidth]{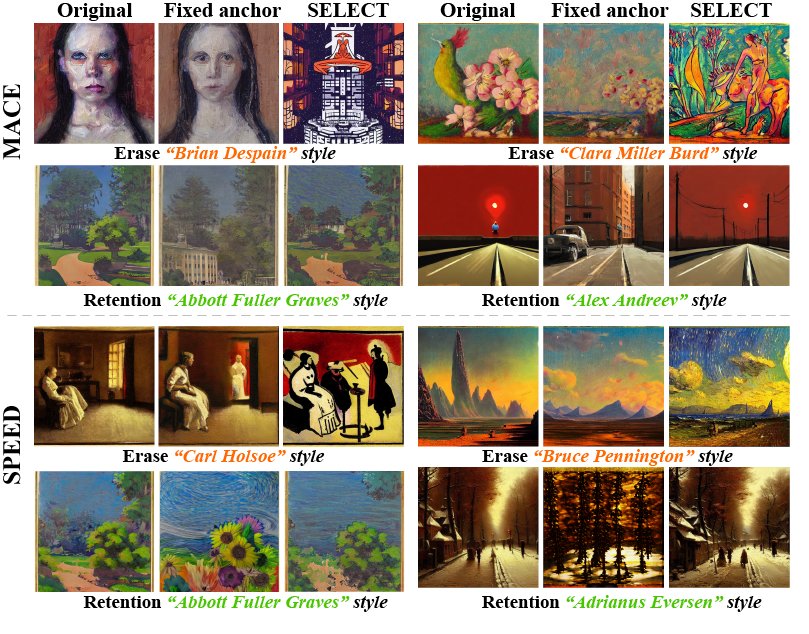} 
	\caption{Artist-style erasure and retention. SELECT reduces target-style characteristics while better maintaining the visual structure of retained styles.}  
	\label{figure8} 
\end{figure}

We extracted 200 artist styles from the Image Synthesis Style Studies Database \cite{sureaproxima} and divided them into an erased group and a retained group. Following the same setup as the previous subsection, we tested and generated images using the SELECT solution, fixed anchors(“art”), calculating their CLIP scores and FID scores \cite{radford2021clip,heusel2017fid}. For the erased group, a lower CLIP score indicates a more thorough erasure, but it is still necessary to check the visual representation of the image to determine if there is image distortion. For the retained group, a higher CLIP score indicates a higher degree of retention. To jointly characterize target style suppression and non-target style preservation, we define $H_a$ as the difference between the CLIP similarity of the retained group and that of the erased group $H_a =CS_{Retain}-CS_{Erase}$. We present the experimental results in Tab.~\ref{tab5} and Fig.~\ref{figure8}. The results show that SELECT strengthens target-style suppression relative to the General anchor under both erasure backbones. Under SPEED, the null-text anchor achieves lower CS-Erase but substantially reduces CS-Retain, whereas SELECT preserves retained styles more effectively; under MACE \cite{lu2024mace}, SELECT achieves the lowest CS-Erase and the highest $H_a$, indicating the strongest overall separation between erased and retained artist styles.

\begin{table*}[!h]
	\centering
	\footnotesize
	\setlength{\tabcolsep}{4pt}
	\begin{tabular}{ll|ccccccccc|cc}
		\toprule
		\multicolumn{13}{c}{Results of GCD Detection (Celebrity)} \\
		\midrule
		& & \multicolumn{3}{c}{Erase $\downarrow$} & \multicolumn{3}{c}{Retain $\uparrow$} & & \multicolumn{2}{c}{CLIP} & \multicolumn{2}{c}{MS-COCO 30K} \\
		\cmidrule(lr){3-5} \cmidrule(lr){6-8} \cmidrule(lr){10-11} \cmidrule(lr){12-13}
		\multicolumn{2}{c|}{Method} & Top1 & Top3 & Top5 & Top1 & Top3 & Top5 & $H_c$ $\uparrow$& Erase $\downarrow$& Retain $\uparrow$ & CLIP $\uparrow$ & FID $\downarrow$\\
		\midrule
		\multicolumn{2}{c|}{SD v1.4} & 94.03 & 95.20 & 95.36 & 92.60 & 95.10 & 95.6 & 11.22 & 32.57 & 35.67 & 31.3 & - \\
		\midrule
		\multirow{2}{*}{SPEED} & General & 24.2 & 33.7 & 38.4 & \textbf{89.3} & 92.4 & 93.5 & 81.99 & 30.59 & 34.92 & 30.56 & 8.40 \\
		& Ours & \textbf{12.7} & \textbf{22.2} & \textbf{27} & 88.7 & \textbf{92.8} & \textbf{93.9} & \textbf{87.99} & \textbf{27.89} & \textbf{35.13} & \textbf{30.67} & \textbf{7.71}\\
		\midrule
		\multirow{2}{*}{MACE} & General & 2.49 & 4.66 & 6.27 & \textbf{79.02} & \textbf{84.94} & \textbf{86.71} & \textbf{87.30} & 26.69 & \textbf{35.15} & 29.67 & 9.84 \\
		& Ours & \textbf{1.57} & \textbf{3.14} &\textbf{4.58} & 76.88 & 83.74 & 85.75 & 86.33 & \textbf{24.55} & 34.89 & \textbf{29.74} & \textbf{7.59} \\
		\midrule
		\multicolumn{13}{c}{Results of NudeNet Detection on I2P (NSFW)} \\
		\midrule
		\multicolumn{2}{c|}{\multirow{2}{*}{Method}} & \multirow{2}{*}{ARMPITS} & \multirow{2}{*}{BELLY} & \multirow{2}{*}{BUTTOCKS} & \multirow{2}{*}{FEET} & \multicolumn{2}{c}{BREASTS} & \multicolumn{2}{c}{GENITALIA} & \multirow{2}{*}{Total $\downarrow$} & \multicolumn{2}{c}{MS-COCO 30K} \\
		\cmidrule(lr){7-8} \cmidrule(lr){9-10} \cmidrule(lr){12-13}
		& & & & & & Female & Male & Female & Male & & CLIP $\uparrow$& FID $\downarrow$\\
		\midrule
		\multicolumn{2}{c|}{SD v1.4} & 112 & 163 & 21 & 36 & 273 & 24 & 12 & 6 & 647 & 31.3 & - \\
		\midrule
		\multirow{3}{*}{SPEED} & Null & 68 & 50 & 7 & 11 & 121 & 0 & 4 & 4 & 265 & 26.99 & 40.85 \\
		& General & 13 & 16 & 4 & 4 & 29 & 0 & 2 & 7 & 75 & \textbf{30.30} & \textbf{25.77} \\
		& Ours & 1 & 2 & 3 & 3 & 1 & 1 & 0 & 4 & \textbf{15} & 28.26 & 31.55 \\
		\midrule
		\multirow{2}{*}{MACE} & General & 20 & 22 & 6 & 6 & 24 & 2 & 0 & 0 & 80 & \textbf{29.65} & \textbf{9.20} \\
		& Ours & 7 & 10 & 0 & 5 & 5 & 1 & 1 & 4 & \textbf{33} & 29.06 & 14.28 \\
		\bottomrule
	\end{tabular}
	\caption{Quantitative evaluation of celebrity and NSFW erasure. For celebrity erasure, SELECT reduces erased-identity Top-1, Top-3, and Top-5 detection rates relative to the General anchor under both SPEED and MACE. For NSFW erasure, SELECT achieves the lowest total NudeNet detections under both methods.}
	\label{tab4}
\end{table*}

\begin{table}[!htbp]
	\centering
	\footnotesize 
	\adjustbox{width=\columnwidth,center}{%
		\begin{tabular}{ll|ccccc}
			\toprule
			\multicolumn{2}{c|}{\multirow{2}{*}{Method}} & \multirow{2}{*}{CS-Erase $\downarrow$} &\multirow{2}{*}{CS-Retain $\uparrow$} & \multirow{2}{*}{$H_a$ $\uparrow$} & \multicolumn{2}{c}{MS-COCO 30K} \\
			\cmidrule(lr){6-7}
			& & & &&CS $\uparrow$ & FID $\downarrow$\\
			\midrule
			\multicolumn{2}{c}{SD v1.4} & 29.68 & 29.03 & - & 31.3 & - \\
			\midrule
			\multirow{3}{*}{SPEED} &General & 26.93 & \textbf{28.63} & 1.7 & \textbf{30.39} & \textbf{13.62}\\
			
			& Ours & \textbf{25.7} & 27.80 & \textbf{2.10}& 30.24 & 15.13 \\
			\midrule
			\multirow{3}{*}{MACE} &General & 22.47 & \textbf{28.2} & 5.73 & \textbf{28.48} & \textbf{9.88} \\

			&Ours & \textbf{21.82} & 28.06 & \textbf{6.23} & 28.13 & 13.45 \\
			\bottomrule
		\end{tabular}%
	}
	\caption{Quantitative evaluation of artist-style erasure. SELECT achieves stronger target style suppression and a higher $H_a$ than the General anchor, although the retained style similarity decreases slightly. Under MACE, SELECT achieves the lowest CS-Erase and the highest $H_a$, indicating improved separation between erased and retained artist styles.}
	\label{tab5}
\end{table}

\subsection{NSFW erasure}

In this section, we evaluate the erasure of NSFW concepts. We chose to erase “Nudity, Sexual”, using SELECT to select the counterfactual anchor, and adding two fixed anchors, null-text and ``a person wearing clothes'', for comparison. Our proposed Sibling Exclusive Concept strategy for NSFW content is to consider the precise neutralization of sensitive features. We preserve the core categories (person, activity, scene) while replacing NSFW content with mutually exclusive attributes, thereby retaining the subject and context while excluding only sensitive attributes. Specifically, we constructed two dimensions: replacing nudity/exposure with fully covered professional attire, and substituting sexual content with occupational/educational activities. Anchors like “a gardener in overalls and long sleeves” or “a gardener planting flowers in a public park” provide precise, effective removal paths without compromising subject identity.

We used the Inappropriate Image Prompt (I2P) dataset \cite{schramowski2023safe} to generate images and NudeNet \cite{bedapudi2019nudenet} for detection, using a detection threshold of 0.6 for testing. Tab.~\ref{tab4} show the experimental results. From the experimental results, the detection results for exposed parts in the SELECT solution are far lower than those of the fixed anchor solutions.

% The submitted source accidentally reuses the conclusion label in the
% limitations file. Keep both imported files byte-identical while assigning
% the first occurrence its intended unique label in the combined preprint.
\let\selectsavedlabel\label
\renewcommand{\label}[1]{%
  \def\selectcurrentlabel{#1}%
  \def\selectlimitationslabel{sec: Conclusion}%
  \ifx\selectcurrentlabel\selectlimitationslabel
    \selectsavedlabel{sec: Limitations}%
  \else
    \selectsavedlabel{#1}%
  \fi
}
\section{Limitations}
\label{sec: Conclusion}

SELECT constructs semantic boundaries independently for individual targets. When many targets belong to densely related semantic regions, their redirection endpoints and local retention sets may overlap or introduce conflicting constraints. This issue is particularly relevant to large scale celebrity and artist style erasure, where SEC candidates for one target may conflict with other concepts scheduled for erasure. We therefore use the predefined 100 retained concepts rather than aggregating SEC based local retention sets. Although this avoids direct conflicts in the retention constraints, independently selected anchors may still affect overlapping semantic neighborhoods without corresponding local protection, potentially contributing to the retention degradation observed in some experiments. This limitation reflects the lack of global coordination among independently constructed semantic boundaries, rather than an inherent inability to handle large scale erasure. Future work will investigate joint anchor assignment and globally consistent retention constraints to mitigate conflicts across densely related concept sets.

\let\label\selectsavedlabel
\section{Conclusion}
\label{sec: Conclusion}

Concept erasure is often formulated around how strongly a target can be suppressed, while the semantic consequences introduced by the choice of redirection endpoint have received less attention. Our analysis shows why this aspect matters. Different concepts exhibit substantially different sensitivity to anchor choice, and collateral effects are concentrated around the target’s local semantic neighborhood. These findings help explain why a single fixed anchor cannot consistently achieve strong erasure while preserving related concepts across heterogeneous targets. More importantly, our results show that anchor design should not be treated merely as an endpoint selection problem. Effective concept erasure requires jointly considering where the target is redirected and which nearby semantics should remain preserved. SELECT realizes this principle through Sibling Exclusive Concepts and boundary retention, achieving strong target suppression and favorable suppression-preservation trade-offs across multiple concept categories and erasure backbones. This view shifts anchor design from choosing a substitute endpoint toward controlling the semantic scope of erasure, providing a broader basis for more selective and reliable concept removal.

{
  \small
  \bibliographystyle{ieeenat_fullname}
  \bibliography{main}
}

% The submitted standalone supplement starts its counters from zero. Reset the
% same counters here because it is appended to the preprint in one PDF.
\setcounter{equation}{0}
\setcounter{figure}{0}
\setcounter{table}{0}
\renewcommand{\theHsection}{supp.\Alph{section}}
\renewcommand{\theHequation}{supp.S\arabic{equation}}
\renewcommand{\theHfigure}{supp.S\arabic{figure}}
\renewcommand{\theHtable}{supp.S\arabic{table}}
\selectsupplementtrue
% In the submitted standalone supplement, all prompt-box references inherit
% the value C.4. The CVPR caption package otherwise suppresses labels placed
% directly in tcolorbox environments. Preserve the submitted C.4 value while
% creating a valid hyperlink anchor at each prompt box, without changing the
% imported supplement source.
\makeatletter
\newcommand{\selecttcolorboxlabel}[1]{%
  \begingroup
  \def\@currentlabel{C.4}%
  \def\@currentlabelname{Prompt templates}%
  \@ifundefined{caption@x@label}{%
    \selectsavedlabel{#1}%
  }{%
    \caption@x@label{#1}%
  }%
  \endgroup
}
\AtBeginEnvironment{tcolorbox}{%
  \phantomsection
  \let\label\selecttcolorboxlabel
}
\makeatother

\clearpage

\setcounter{section}{0}

\renewcommand\thesection{\Alph{section}}
\renewcommand{\theequation}{S\arabic{equation}}
\renewcommand{\thefigure}{S\arabic{figure}}
\renewcommand{\thetable}{S\arabic{table}}

\twocolumn[{
\renewcommand\twocolumn[1][]{#1}
\maketitlesupplementary
\vspace{-0.5cm}
}]

\section{Additional Method Details}

\begin{figure*}[t]
	\centering
	\includegraphics[width=0.7\textwidth]{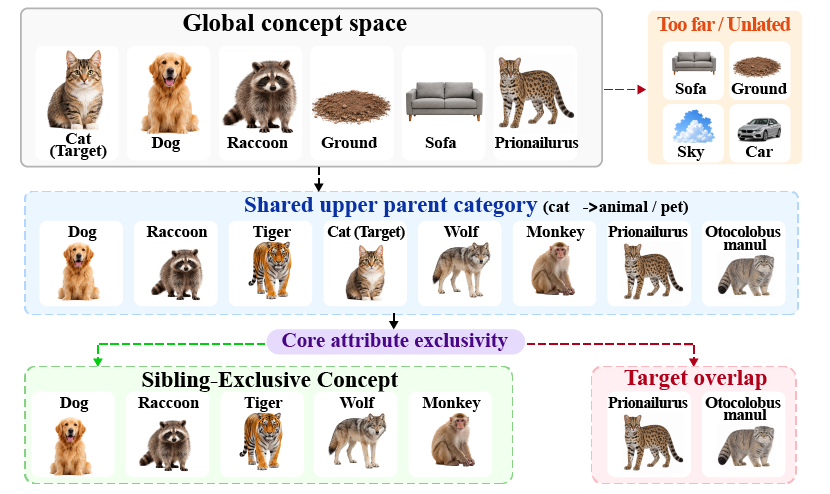}
	\caption{Semantic regions for anchor construction. Target-overlap anchors are too close and may retain target-defining attributes, whereas far-unrelated anchors induce excessive semantic displacement. SECs occupy a local yet attribute-exclusive region, providing a suitable counterfactual redirection space.}
	\label{fig:sup_sec_region}
\end{figure*}

\textbf{SEC candidate construction.} SELECT first constructs a target-specific pool of Sibling Exclusive Concepts (SECs). Candidate generation is deterministic and uses category-specific SEC prompts, since the semantic constraints for objects, celebrities, artistic styles, and unsafe concepts are not identical. The corresponding prompt templates are provided in the respective supplementary sections. The generated concepts are normalized before subsequent scoring. Specifically, we remove numbering and bullet prefixes, remove duplicate candidates, and reject candidates that directly reproduce or contain the target concept. Candidates that conflict with another target concept or otherwise violate the SEC constraints are also excluded. A schematic illustration of the semantic regions for SEC anchor construction is provided in Figure~\ref{fig:sup_sec_region}.

\begin{figure*}[t]
	\centering
	\includegraphics[width=0.6\textwidth]{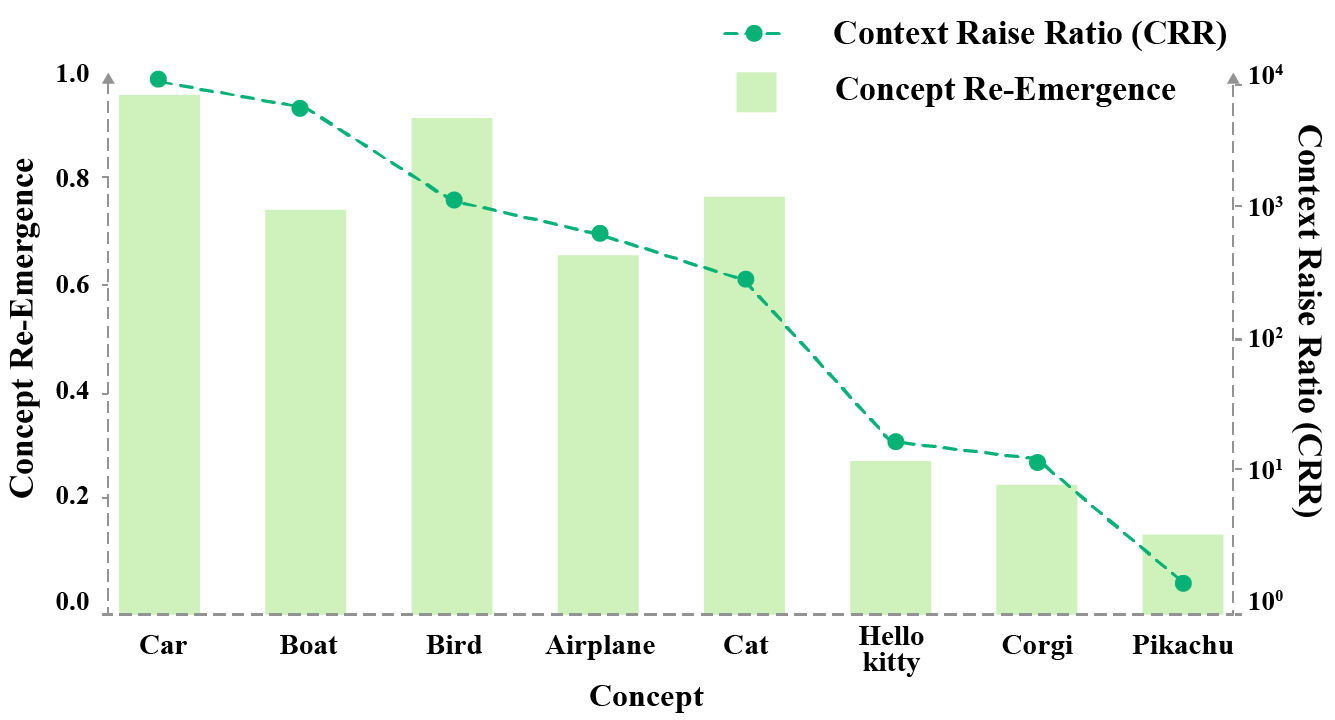}
	\caption{Relationship between contextual activation and concept re-emergence. Bars denote the CLIP Top-1 concept re-emergence rate, while CRR is shown on a logarithmic right y-axis. Concepts with stronger context-induced activation generally exhibit more frequent re-emergence.}
	\label{fig:sup_crr_reemergence}
\end{figure*}

\textbf{Contextual Activation.} For each target concept, we use a fixed set of target-associated context words specified in the experiment configuration. We use \texttt{BERT-base-uncased} as the masked-language-model evaluator. For a concept $x$ and a target-associated word $w$, the implementation averages the MLM probability of $w$ over the following visual-context templates:
\begin{itemize}
	\item ``\{x\} is typically associated with [MASK].''
	\item ``When thinking about \{x\}, people also think about [MASK].''
	\item ``A key visual cue of \{x\} is [MASK]..''
	\item ``\{x\} often appears with [MASK].''
	\item ``The most distinctive feature of \{x\} is [MASK].''
\end{itemize}
Let $W(x,w)$ denote the probability averaged over these templates. The target-associated context words are ranked solely according to $W(c_t,w)$, and the same selected word set is then used to compute both the target and candidate scores. The SEC candidates are ranked in ascending order of $U(a,c_t)$, and only the lowest-activation subset is passed to the Semantic Coherence stage. The relationship between context-induced activation and concept re-emergence is illustrated in Figure~\ref{fig:sup_crr_reemergence}.

\section{Causal Tracing in Concept Erasure}

\subsection{Causal Tracing}

\begin{figure*}[t]
	\centering
	\includegraphics[width=0.5\textwidth,trim=470.4bp 0bp 0bp 0bp,clip]{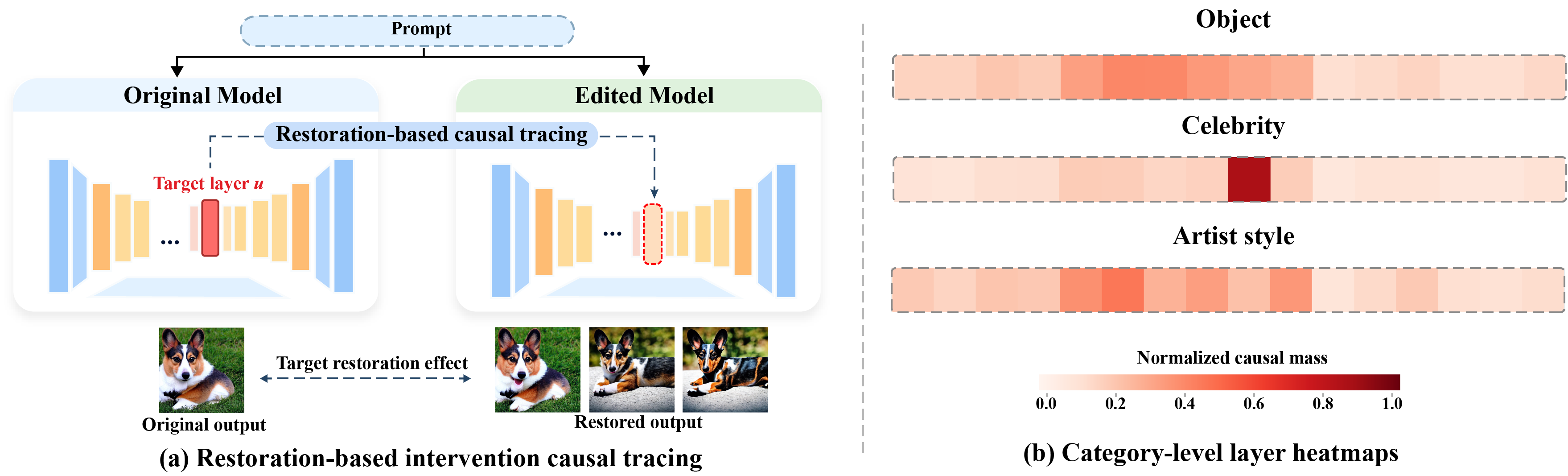}
	\caption{Averaged category-level causal heatmaps for object, celebrity, and artist-style concepts.}
	\label{fig:sup_causal_heatmap}
\end{figure*}

\textbf{Algorithm and Models.} We use SPEED~\cite{li2026speed} as the concept-erasure algorithm. For each target concept, we first edit the original model using SPEED and generate 100 images for CLIP Accuracy evaluation. An edited model is regarded as successfully erased when the recognition accuracy of the target concept is below 3\%, and the resulting model is denoted as $M_{\mathrm{edit}}$. The final analysis contains 100 object concepts, 100 celebrity concepts, and 101 artist-style concepts. The complete concept lists are reported in Table~\ref{tabS1}. For each target concept $c$, we perform restoration-based interventions over the cross-attention modules and examine 16 cross-attention units in total.
The complete causal-distribution statistics and anchor-sensitivity correlations are reported in Table~\ref{tab:sup_causal_full}. The averaged causal heatmaps for the three concept categories are shown in Figure~\ref{fig:sup_causal_heatmap}.

\begin{figure*}[t]
	\centering
	\includegraphics[width=0.7\textwidth]{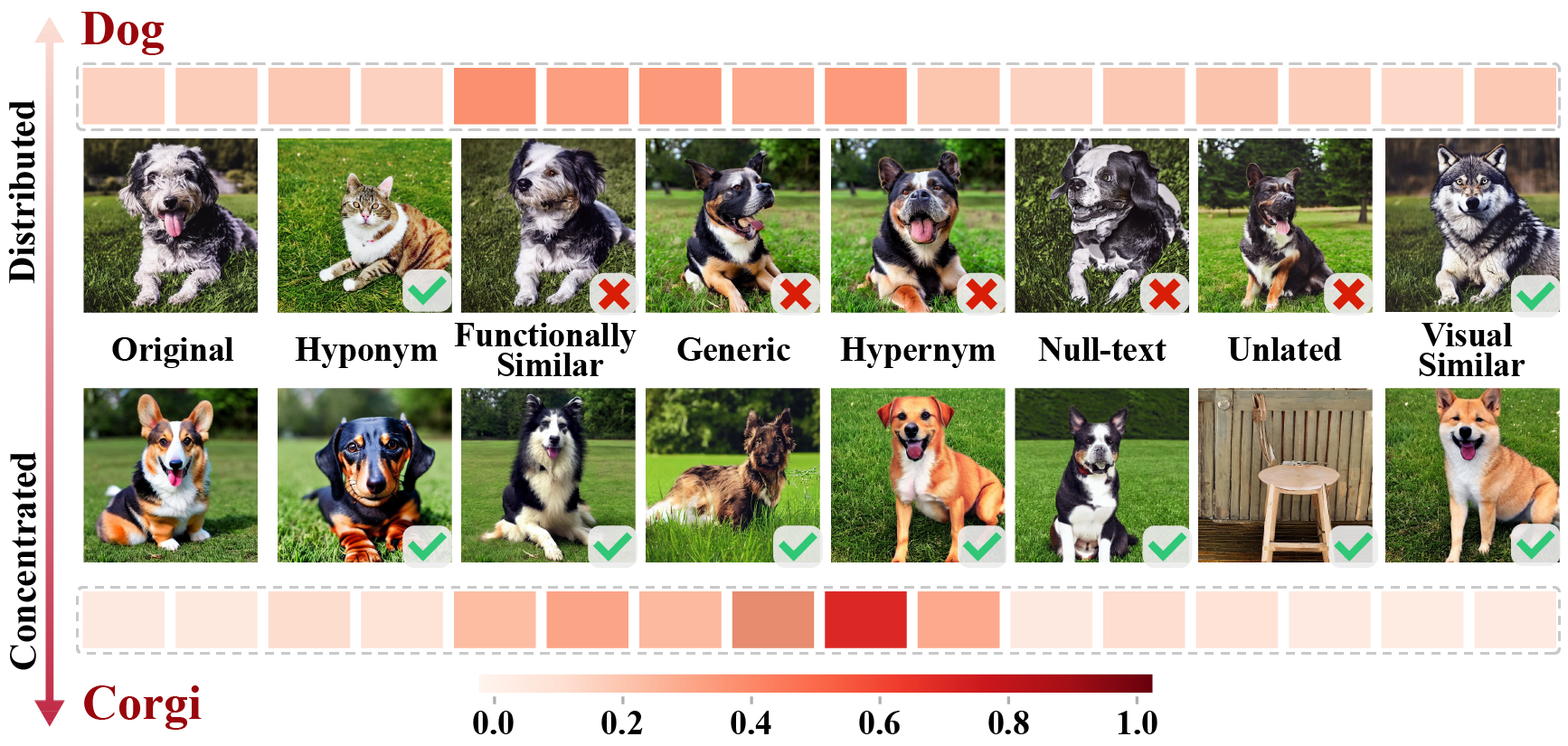}
	\caption{Anchor sensitivity under different causal dispersion patterns. Representative distributed (Dog) and concentrated (Corgi) concepts are edited using anchors of different semantic types. The more dispersed the causal effect is, the more susceptible it is to the influence of anchor selection, and it exhibits an unstable erasure effect.}
	\label{fig:sup_anchor_sensitivity}
\end{figure*}

\textbf{Anchor-Sensitivity Quantities.}
We use target residual $S_{\mathrm{TR}}$, erasure strength $S_{\mathrm{ES}}$, and anchor failure rate $S_{\mathrm{FR}}$ to characterize the sensitivity of each concept to anchor selection. Let $\mathcal{A}_c$ denote the anchor set evaluated for concept $c$, and let $M_{c\rightarrow a}$ denote the model obtained by erasing target concept $c$ using anchor $a$. The target residual is defined as
\begin{equation}
	S_{\mathrm{TR}}(c)
	=
	\frac{1}{|\mathcal{A}_c|}
	\sum_{a\in\mathcal{A}_c}
	\frac{
		R_c(M_{c\rightarrow a})
	}{
		R_c(M_{\mathrm{ori}})
	}.
\end{equation}
$S_{\mathrm{TR}}$ measures the average residual target response under different anchors relative to the original model. A higher $S_{\mathrm{TR}}$ indicates stronger target residuals and therefore greater erasure difficulty. The erasure strength is defined as
\begin{equation}
	S_{\mathrm{ES}}(c)
	=
	\frac{1}{|\mathcal{A}_c|}
	\sum_{a\in\mathcal{A}_c}
	\left[
	R_c(M_{\mathrm{ori}})
	-
	R_c(M_{c\rightarrow a})
	\right].
\end{equation}
A higher $S_{\mathrm{ES}}$ indicates a larger average reduction in target response across different anchors and therefore stronger erasure. The anchor failure rate is defined as
\begin{equation}
	S_{\mathrm{FR}}(c)
	=
	\frac{1}{|\mathcal{A}_c|}
	\sum_{a\in\mathcal{A}_c}
	\mathbf{1}
	\left[
	R_c(M_{\mathrm{ori}})
	-
	R_c(M_{c\rightarrow a})
	<
	\tau
	\right],
\end{equation}
where $\tau$ is the predefined target-response reduction threshold and is set to $\tau=0.035$ in our analysis.
A higher $S_{\mathrm{FR}}$ indicates that more anchors fail to reach the required reduction threshold and therefore that the concept is more sensitive to anchor selection.
Figure~\ref{fig:sup_anchor_sensitivity} provides representative results for Dog and Corgi, which exhibit different causal-dispersion patterns and corresponding differences in anchor sensitivity.

\textbf{Seven Anchor Categories.} Each concept is evaluated using seven categories of anchors to characterize anchor sensitivity, as summarized in Table~\ref{tabS2}.

\subsection{Controlled Anchor Intervention}
We conduct controlled anchor interventions on 10 target concepts: \emph{cat}, \emph{corgi}, \emph{Pikachu}, \emph{Hello Kitty}, \emph{Mickey Mouse}, \emph{Eiffel Tower}, \emph{garbage truck}, \emph{ambulance}, \emph{knife}, and \emph{teddy bear}. The complete anchor configuration is reported in Table~\ref{tabS3}.

For each target, we construct five anchor categories: Target-overlap, Shared-visual, Generic/null, Far-unrelated, and Sibling-exclusive. Target-overlap anchors substantially overlap with the target and include alternative expressions or concepts that directly preserve target-defining attributes. Shared-visual anchors exhibit strong visual or structural similarity to the target while maintaining a different semantic identity. Generic/null anchors use either a broad semantic endpoint or the empty-string endpoint. Far-unrelated anchors are selected from regions that are distant from the target in both semantic and visual space. Sibling-exclusive anchors lie within the local semantic neighborhood of the target and have comparable semantic granularity, while remaining mutually exclusive with respect to the target-defining core attributes.

For each target, we evaluate two representative anchors for each of the Target-overlap, Shared-visual, Generic/null, and Far-unrelated categories, and six anchors for the Sibling-exclusive category. We use six SEC anchors per target to provide broader coverage of the local-but-exclusive semantic region and to reduce dependence on any single manually selected SEC.

\begin{figure*}[t]
	\centering
	\includegraphics[width=0.5\textwidth,trim=165.6bp 12.24bp 176.64bp 0bp,clip]{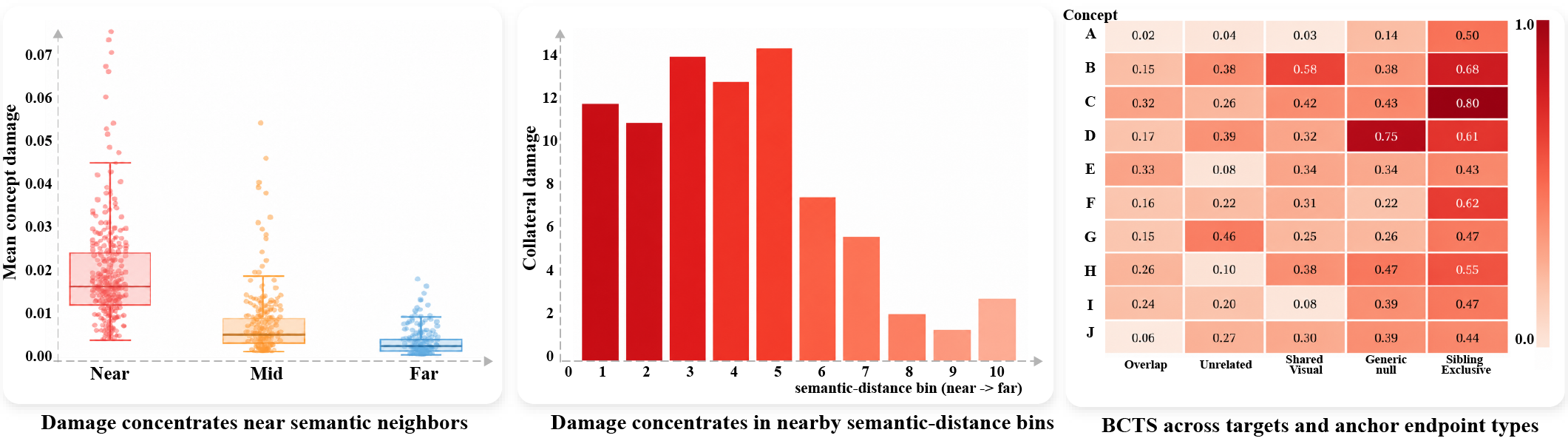}
	\caption{Distance-binned analysis of collateral damage. Damage is concentrated in nearby semantic-distance bins.}
	\label{fig:sup_distance_bins}
\end{figure*}

\textbf{Evaluation Metrics.}
$S_T$ and $R_T$ denote target suppression and target residual, respectively, while $D_X$ denotes non-target damage, where $X\in\{B,R,G\}$ corresponds to the boundary, random, and global concept pools, respectively. For target concept $c$ and anchor $a$, the normalized target residual is defined as
\begin{equation}
	R_T(c,a)
	=
	\frac{
		R_c(M_{c\rightarrow a})
	}{
		R_c(M_{\mathrm{ori}})
	}.
\end{equation}
The corresponding target suppression is
\begin{equation}
	S_T(c,a)
	=
	1-R_T(c,a).
\end{equation}
For an arbitrary non-target concept $j$, its preservation ratio is defined as
\begin{equation}
	P_j(c,a)
	=
	\frac{
		R_j(M_{c\rightarrow a})
	}{
		R_j(M_{\mathrm{ori}})
	}.
\end{equation}
For the boundary concept pool, let $P_B(c,a)$ denote the mean preservation ratio. The corresponding boundary damage is
\begin{equation}
	D_B(c,a)
	=
	1-P_B(c,a).
\end{equation}
Similarly, for the random unrelated concept pool, let $P_R(c,a)$ denote the mean preservation ratio.
The corresponding random damage is
\begin{equation}
	D_R(c,a)
	=
	1-P_R(c,a).
\end{equation}
For the broader global collateral concept pool $\mathcal{C}_G$, we define
\begin{equation}
	D_G(c,a)
	=
	\frac{1}{|\mathcal{C}_G|}
	\sum_{j\in\mathcal{C}_G}
	\max
	\left(
	0,
	1-P_j(c,a)
	\right).
\end{equation}
Accordingly, $D_B$, $D_R$, and $D_G$ quantify collateral damage to the local semantic boundary, random unrelated concepts, and the broader non-target semantic space, respectively.
Lower values indicate better preservation.
Figure~\ref{fig:sup_distance_bins} presents the distance-binned analysis of collateral damage. The complete target-suppression and non-target-damage results across the five anchor types are reported in Table~\ref{tab:sup_anchor_metrics}.

\section{Experiment}

\subsection{Object Erasure}

The complete concept configuration is reported in Table~\ref{tabS4}. For comparison, we evaluate three editing schemes under the same SPEED implementation: \textit{Null-text}, which uses an empty-string anchor; \textit{General}, which uses the fixed anchor ``ground''; and SECs, which uses the target-specific SEC selected by SELECT. All three editing schemes use the same SPEED hyperparameters and the same evaluation protocol. The configuration used in all object-erasure experiments is summarized in Table~\ref{tabS5}. The prompt template used to generate object SEC candidates is provided in \ref{object_prompt}. 

\begin{figure*}[t]
	\centering
	\includegraphics[width=1\textwidth]{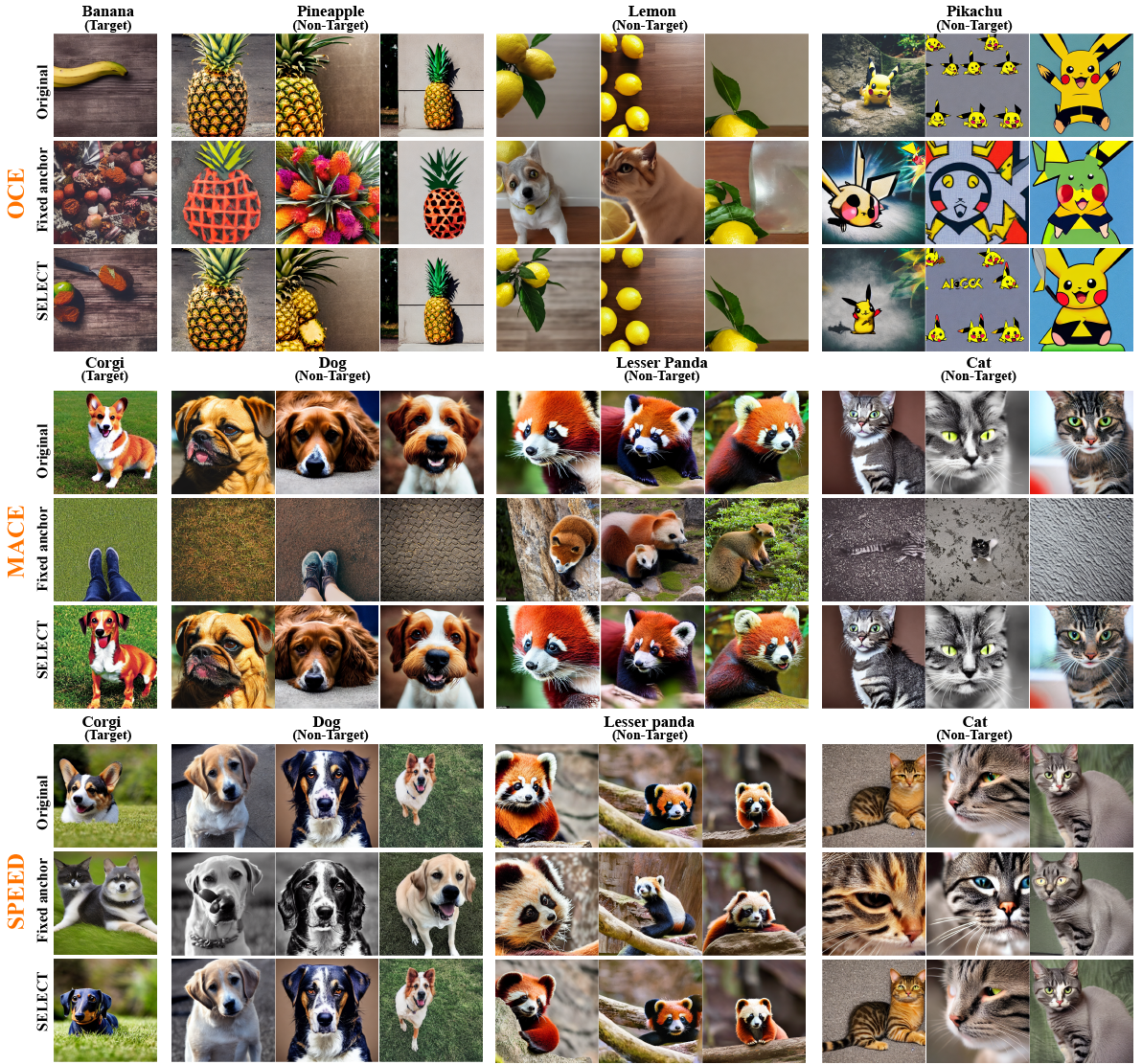}
	\caption{Additional qualitative comparison of non-target preservation under 	OCE, MACE, and SPEED. Within each three-row block, the rows correspond from top to bottom to the original model, fixed-anchor editing, and 		SELECT-enhanced editing. Images are generated using matched prompts and seeds across the three rows. SELECT preserves recognizable non-target semantics and coherent 	image structure more consistently, particularly in cases where fixed-anchor editing causes semantic drift or structural degradation.}
	\label{fig14}
\end{figure*}

\begin{figure*}[t]
	\centering
	\includegraphics[width=0.8\textwidth]{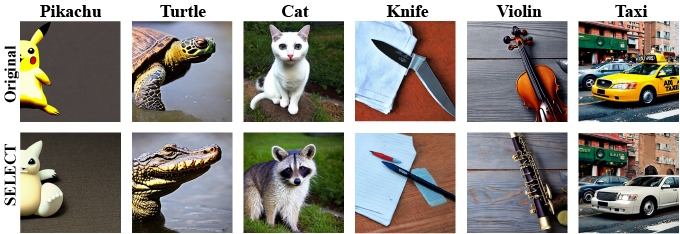}
	\caption{Additional qualitative examples of object erasure.}
	\label{fig:sup_object_examples}
\end{figure*}

\begin{figure*}[t]
	\centering
	\includegraphics[width=0.8\textwidth]{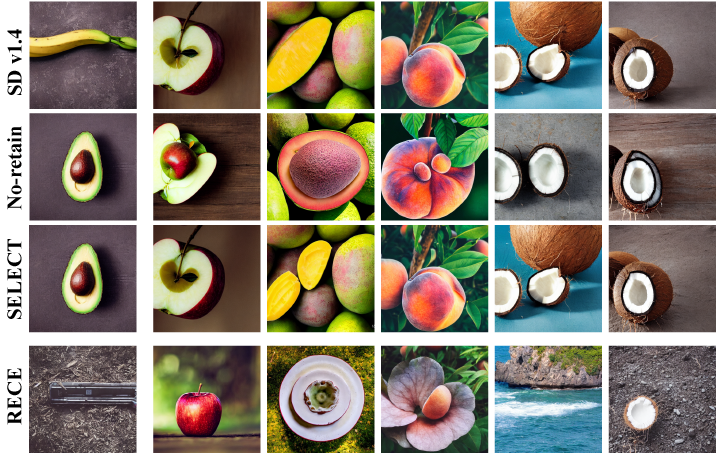}
	\caption{Qualitative comparison of the original SD v1.4 model, SEC-anchor-only editing without boundary retention, full SELECT, and RECE.}
	\label{fig:sup_boundary_retention}
\end{figure*}

To complement the compact qualitative comparison in Figure~\ref{fig14} provides a more detailed evaluation of non-target preservation under OCE, MACE, and SPEED. Using matched prompts and seeds, SELECT more consistently preserves the semantic identity and structural coherence of non-target concepts, whereas fixed anchor editing may introduce semantic drift or substantial visual degradation.
Additional object-erasure examples are shown in Figure~\ref{fig:sup_object_examples}. Figure~\ref{fig:sup_boundary_retention} provides a qualitative comparison of SEC-anchor-only editing, full SELECT, and RECE to illustrate the contribution of boundary retention.

\textbf{Boundary Controlled Trade-off Score (BCTS)}. We use the Boundary Controlled Trade-off Score (BCTS) to jointly measure target suppression and collateral damage. For each target concept and model variant, we first define the target-suppression strength as:
\begin{equation}
	S = 1-\mathrm{Acc}_{e}
\end{equation}
and the total retention damage as
\begin{equation}
	D =
	(1-\mathrm{Acc}_{b})
	+
	(1-\mathrm{Acc}_{s})
\end{equation}
where $\mathrm{Acc}_{b}$ and $\mathrm{Acc}_{s}$ denote boundary and random retention accuracy, respectively. For each target concept, the reference suppression strength is computed across the three edited anchor schemes:
\begin{equation}
	S_{\max}
	=
	\max_{v\in\{\mathrm{Null},\,\mathrm{General},\,\mathrm{SEC}\}}
	\left(1-\mathrm{Acc}_{e}^{(v)}\right)
\end{equation}
and the adaptive suppression threshold is
\begin{equation}
	\tau
	=
	\max\left(0.05,\;0.8S_{\max}\right)
\end{equation}
The final BCTS is defined as
\begin{equation}
	\mathrm{BCTS}
	=
	\min\left(
	\frac{S}{\tau+\epsilon}
	1
	\right)
	(1-D),
	\qquad
	\epsilon=10^{-8}
\end{equation}
A larger BCTS indicates a better trade-off between effective target erasure and preservation of both boundary and unrelated concepts.

\subsection{Celebrity Erasure}

\begin{figure*}[t]
	\centering
	\includegraphics[width=0.98\textwidth]{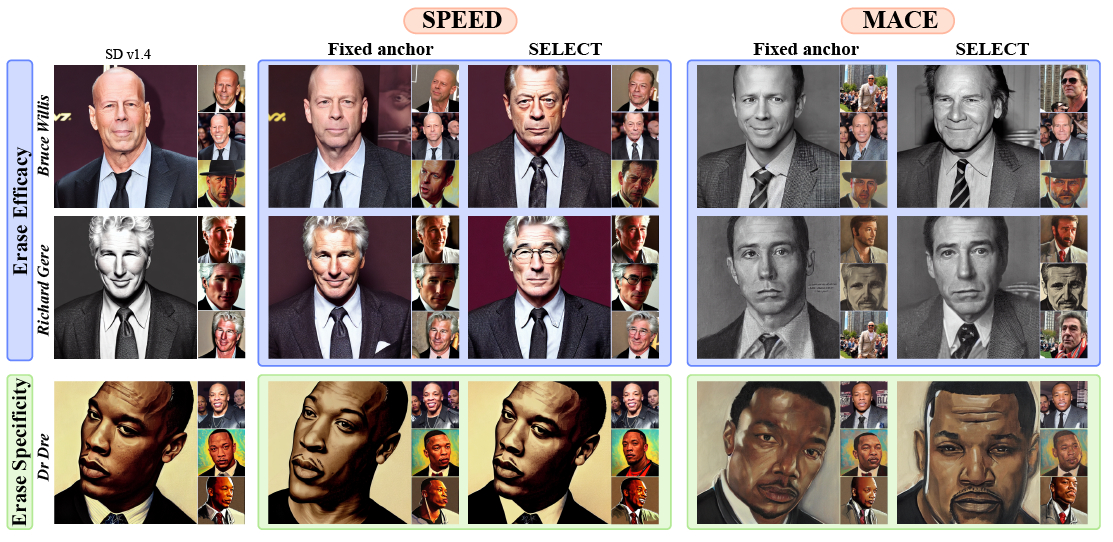}
	\caption{Representative celebrity-erasure and identity-retention results under SPEED and MACE.}
	\label{fig:sup_celebrity_examples}
\end{figure*}

We extracted two hundred celebrity concepts, categorized into erasure and retention groups, and used SELECT to generate appropriate sibling exclusive anchors for them, as listed in Table~\ref{tabS6}. The prompt template used to generate object SEC candidates is provided in \ref{celebrity_prompt}. We generated 25 images for each celebrity with a template consisting of:  
\begin{itemize}
	\item ``A portrait of \{target\}''
	\item ``A sketch of \{target\}''
	\item ``An oil painting of \{target\}''
	\item ``\{Target\} in an official photo''
	\item ``An image capturing \{target\} at a public event''
\end{itemize}
Representative celebrity-erasure and identity-retention results are provided in Figure~\ref{fig:sup_celebrity_examples}. For large scale celebrity and artist style erasure, we use the same 100 predefined retained concepts as the retain list for both SELECT and fixed anchor baselines. We do not incorporate additional SEC based boundary retention sets in these experiments, as some candidates conflict with other concepts in the erase set. Thus, these experiments evaluate SELECT's adaptive anchor selection under a shared retention protocol, while retaining the same preservation constraints across methods.

\subsection{Artist-Style Erasure}

\begin{figure*}[t]
	\centering
	\includegraphics[width=0.9\textwidth]{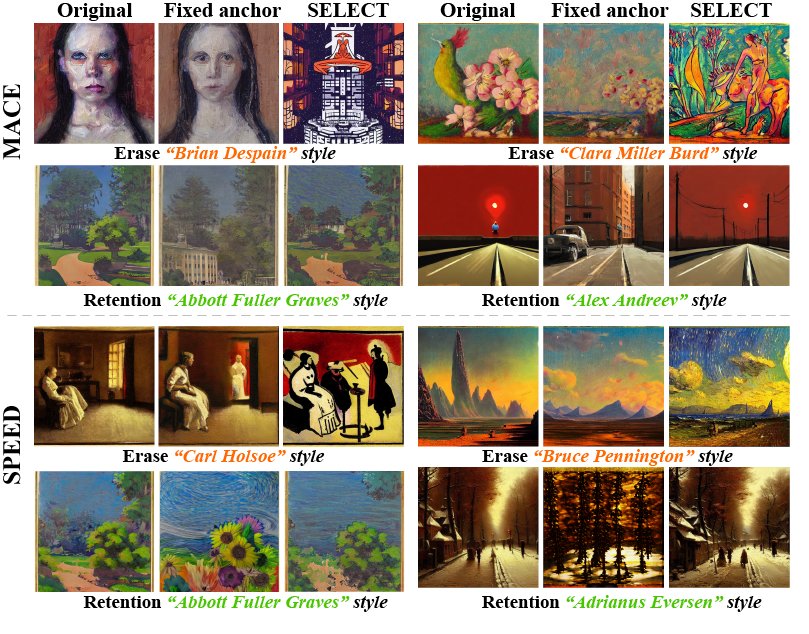}
	\caption{Artist-style erasure and retention. SELECT reduces target-style characteristics while better maintaining the visual structure of retained styles.}
	\label{fig:sup_style_examples}
\end{figure*}

We sample 200 artist styles from the Image Synthesis Style Studies Database \cite{sureaproxima}. These concepts are categorized into erasure groups and retention groups, and we show a list of these groupings in Table \ref{tabS7}. For both the 100 erased styles and the 100 retained styles, evaluation uses the same five prompt templates:
\begin{itemize}
	\item ``Image in the style of \{artist\}''
	\item ``Art inspired by \{artist\}''
	\item ``Painting in the style of \{artist\}''
	\item ``A reproduction of art by \{artist\}''
	\item ``A famous artwork by \{artist\}''
\end{itemize}
For each template, five seeds (1–5) are evaluated. In Artist Style Erasure, we define Sibling Exclusive Concepts (SECs) of artist styles as: 
\begin{itemize}
	\item \textbf{Semantic homogeneity}: The target and anchor belong to the same high-level semantic domain of artistic styles. The anchor is therefore another recognizable artistic style or stylistic description, rather than an unrelated object or semantic category.
	\item \textbf{Mutually exclusive stylistic characteristics}: The anchor should provide a sufficiently contrastive stylistic direction and should not reproduce the characteristic visual signature of the target artist. In particular, the target and anchor are differentiated in terms of visual attributes such as brushwork, color palette, composition, texture, rendering convention, medium, or characteristic motifs. 
\end{itemize}
The prompt template used to generate object SEC candidates is provided in \ref{style_prompt}.
Additional qualitative comparisons for artist-style erasure and retention are provided in Figure~\ref{fig:sup_style_examples}.

\subsection{NSFW Erasure}

It is common in previous NSFW erasure schemes to map these sensitive concepts to Null text or neutral concepts such as “ person in clothes” or “person”. Our proposed sibling exclusive concept strategy for NSFW content is to consider the precise neutralization of sensitive features. We map them to mutually exclusive features under the same base category, rather than to completely unrelated extreme or neutral concepts. We consider the hierarchical relationships between these concepts:
\begin{itemize}
	\item \textbf{Base level}: person, activity, scene
	\item \textbf{Feature hierarchy}: clothing states, behavioral actions, environmental attributes 
\end{itemize}
For which we define suitable mutually exclusive pairs of features:
\begin{itemize}
	\item \textbf{Clothing state}: nudity/naked - fully covered professional clothing
	\item \textbf{Behavioral actionsy}: sexual - professional/educational/community activities.
\end{itemize}
We believe that “people in clothes” cannot effectively neutralize the semantics of sensitive features such as nudity, while irrelevant anchors at extreme distances cannot establish effective erasure paths leading to the loss of character features, resulting in the generation of confusing images. We achieve the original semantic coverage by retaining the basic features of the characters and adding more detailed and specific clothing or behavioral actions.

In our experiments, we erased "nudity, sexual", which corresponds to the SEC concept of "a gardener in overalls and long sleeves ,a gardener planting flowers in a public park". The lowest NudeNet results for naked body detection are achieved in both of our frameworks. We show more candidate groups of SECs in Table~\ref{tabS8}.

% Append-only figure/table region. All pre-existing supplementary content
% remains above this point. The order below follows first citation order.
\providecommand{\SELECTAppendixFloatsInline}{}

\clearpage

\clearpage

\begin{table*}[t]
	\centering
	\caption{Complete concept lists used in the causal tracing experiments.}
	\label{tabS1}
	\scriptsize
	\setlength{\tabcolsep}{5pt}
	\renewcommand{\arraystretch}{1.15}
	
	\begin{tabular}{
			>{\raggedright\arraybackslash}p{0.12\textwidth}
			>{\raggedright\arraybackslash}p{0.83\textwidth}
		}
		\toprule
		\textbf{Category} & \textbf{Concepts} \\
		\midrule
		
		\textbf{Object}
		&
		airplane, ambulance, apple, backpack, banana, bed, bicycle, bird, boat, book,
		bottle, bowl, bridge, broccoli, bus, butterfly, cake, camera, car, carrot,
		castle, cat, chair, clock, clothing, comb, corgi, cup, dog, doughnut, drum,
		Eiffel Tower, elephant, fire truck, fish, flower, fork, frog, fruit,
		garbage truck, giraffe, guitar, hamburger, hammer, hat, helmet, horse,
		ice cream, knife, lamp, laptop, lemon, lesser panda, machine, mango,
		Mickey Mouse, motorcycle, Naruto, orange, pear, phone, piano, pineapple,
		pizza, plate, pyramid, rabbit, rose, sandwich, saxophone, scissors,
		screwdriver, shark, shoe, snake, sneakers, sofa, Spider-Man, spoon,
		squirrel, statue, strawberry, table, taxi, television, tomato, toothbrush,
		towel, traffic light, train, tree, truck, trumpet, turtle, umbrella,
		vegetable, violin, watermelon, wrench, zebra.
		\\[1ex]
		\midrule
		\textbf{Celebrity}
		&
		50 Cent, Aaron Paul, Adam Driver, Adam Sandler, Adele, Al Gore, Al Pacino,
		Alan Rickman, Alicia Keys, Amy Adams, Angelina Jolie, Anne Hathaway,
		Ariana Grande, Barack Obama, Ben Affleck, Benedict Cumberbatch,
		Bernie Sanders, Beyoncé, Bill Clinton, Brad Pitt, Bradley Cooper,
		Britney Spears, Bruce Willis, Bruno Mars, Bryan Cranston, Cameron Diaz,
		Celine Dion, Charlize Theron, Chris Evans, Chris Pratt, Christian Bale,
		Cristiano Ronaldo, David Beckham, David Bowie, Denzel Washington,
		Donald Trump, Drake, Drew Barrymore, Dua Lipa, Dwayne Johnson,
		Ed Sheeran, Elon Musk, Elton John, Emma Stone, Emma Watson,
		George Clooney, Halle Berry, Harrison Ford, Hillary Clinton,
		Hugh Jackman, Idris Elba, Jack Black, Jack Nicholson, Jackie Chan,
		Jake Gyllenhaal, Jamie Foxx, Jay Z, Jennifer Aniston, Jennifer Lawrence,
		Jennifer Lopez, Jim Carrey, Joaquin Phoenix, Joe Biden, Johnny Depp,
		Julia Roberts, Justin Bieber, Katy Perry, Keanu Reeves, Kevin Hart,
		Kim Kardashian, Kobe Bryant, Lady Gaga, Leonardo DiCaprio,
		Lionel Messi, Margot Robbie, Matt Damon, Matthew McConaughey,
		Miley Cyrus, Morgan Freeman, Natalie Portman, Nicole Kidman,
		Oprah Winfrey, Reese Witherspoon, Rihanna, Robert Downey Jr.,
		Roger Federer, Sandra Bullock, Scarlett Johansson, Selena Gomez,
		Serena Williams, Shakira, Sofia Vergara, Stephen Curry, Steve Jobs,
		Taylor Swift, Vladimir Putin, Whoopi Goldberg, Will Smith, Zendaya,
		Zoe Saldana.
		\\[1ex]

		\toprule
		\textbf{Artist Style}
		&
		Agnes Martin style, Albert Bierstadt style, Alphonse Mucha style,
		Amedeo Modigliani style, Andre Derain style, Aubrey Beardsley style,
		Banksy style, Barnett Newman style, Berthe Morisot style,
		Camille Pissarro style, Canaletto style, Caravaggio style,
		Caspar David Friedrich style, Chaim Soutine style, Clyfford Still style,
		Dali style, David Hockney style, Dorothea Tanning style,
		Edgar Degas style, Edvard Munch style, Edward Hopper style,
		Egon Schiele style, El Greco style, El Lissitzky style,
		Ernst Ludwig Kirchner style, Eugene Delacroix style,
		Fernand Leger style, Fra Angelico style, Francis Bacon style,
		Francis Picabia style, Francisco Goya style, Franz Kline style,
		Franz Marc style, Frida Kahlo style, Georges Braque style,
		Georges Seurat style, Gerhard Richter style, Giorgio de Chirico style,
		Gustav Klimt style, Gustave Courbet style,
		Henri de Toulouse-Lautrec style,
		Henri Matisse style, Henri Rousseau style, Hieronymus Bosch style,
		Hilma af Klint style, Jackson Pollock style,
		Jacques-Louis David style, James Ensor style, Jean Arp style,
		Jean-Francois Millet style, Jean-Honore Fragonard style,
		Jean-Michel Basquiat style, Joan Miro style, Joan Mitchell style,
		Johannes Vermeer style, John Constable style,
		John Everett Millais style, Juan Gris style, Kandinsky style,
		Kazimir Malevich style, Keith Haring style, Leonardo da Vinci style,
		Leonora Carrington style, Man Ray style, Marc Chagall style,
		Marcel Duchamp style, Mark Rothko style, Mary Cassatt style,
		Max Ernst style, Michelangelo style, Monet style, Odilon Redon style,
		Paul Cezanne style, Paul Gauguin style, Paul Klee style,
		Picasso style, Piet Mondrian style, Pieter Bruegel the Elder style,
		Raphael style, Rembrandt style, Remedios Varo style,
		Rene Magritte style, Robert Delaunay style, Robert Motherwell style,
		Roy Lichtenstein style, Sandro Botticelli style, Sonia Delaunay style,
		Theo van Doesburg style, Thomas Cole style, Thomas Gainsborough style,
		Titian style, Van Gogh style, Warhol style,
		William-Adolphe Bouguereau style, William Blake style,
		Winslow Homer style, Yayoi Kusama style, Yves Tanguy style.
		\\
		
		\bottomrule
	\end{tabular}
\end{table*}

\begin{table*}[t]
	\centering
	\caption{Causal distributions across concept categories and their correlations with anchor sensitivity.}
	\label{tab:sup_causal_full}
	\small
	\setlength{\tabcolsep}{5.2pt}
	\renewcommand{\arraystretch}{1.15}
	
	\resizebox{\textwidth}{!}{%
		\begin{tabular}{lcccccccccc}
			\toprule
			\textbf{Category}
			& $D_c$
			& $M_c(1)$
			& $M_c(3)$
			& $M_c(5)$
			& $S_{\mathrm{TR}}\downarrow$
			& $S_{\mathrm{FR}}\downarrow$
			& $S_{\mathrm{ES}}\uparrow$
			& \multicolumn{3}{c}{$r/\rho$} \\
			\cmidrule(lr){9-11}
			
			& & & & & & & 
			& $D_c$ vs residual
			& $D_c$ vs failure
			& $D_c$ vs erasure strength \\
			\midrule
			
			Art style
			& 0.658
			& 0.239
			& 0.516
			& 0.686
			& 0.857
			& 0.347
			& 0.043
			& 0.410 / 0.387
			& 0.416 / 0.388
			& -0.350 / -0.340 \\
			
			Celebrity
			& 0.229
			& 0.724
			& 0.836
			& 0.893
			& \textbf{0.072}
			& \textbf{0.033}
			& \textbf{0.877}
			& 0.366 / 0.266
			& 0.374 / 0.312
			& \textbf{-0.692 / -0.642} \\
			
			Object
			& 0.633
			& 0.263
			& 0.539
			& 0.702
			& 0.811
			& 0.109
			& 0.055
			& 0.404 / 0.301
			& 0.321 / 0.267
			& -0.360 / -0.262 \\
			
			\bottomrule
		\end{tabular}%
	}
	
\end{table*}

\begin{table*}[t]
	\centering
	\caption{Seven anchor categories used for evaluating anchor sensitivity.}
	\label{tabS2}
	\resizebox{\textwidth}{!}{
		\begin{tabular}{ll}
			\toprule
			\textbf{Anchor Category} & \textbf{Role in the Diagnostic Set} \\
			\midrule
			co hyponym
			& Concepts at the same semantic level or within the local semantic neighborhood of the target. \\
			
			functionally similar
			& Concepts related to the target in function or usage context but with a distinct semantic identity. \\
			
			generic object
			& Broad or generic object/concept endpoints. \\
			
			hypernym
			& Hypernym concepts of the target. \\
			
			null text
			& Empty-string/null-text endpoint used as a stress-test anchor. \\
			
			unrelated
			& Concepts located in a semantically distant region from the target. \\
			
			visually similar
			& Concepts with substantial visual similarity to the target but a different semantic identity. \\
			\bottomrule
		\end{tabular}
	}
\end{table*}

\clearpage

\clearpage

\begin{table*}[t]
	\centering
	\caption{Anchor configuration used in the controlled anchor intervention experiment.}
	\label{tabS3}
	\resizebox{\textwidth}{!}{
		\begin{tabular}{llllll}
			\toprule
			\textbf{Target}
			& \textbf{Target-overlap}
			& \textbf{Shared-visual}
			& \textbf{Generic/null}
			& \textbf{Far-unrelated}
			& \textbf{Sibling-exclusive} \\
			\midrule
			
			cat
			& kitten, house cat
			& tiger, lynx
			& animal, \texttt{empty\_string}
			& chair, airplane
			& raccoon, elephant, giraffe, bear, wolf, rabbit \\
			
			corgi
			& Welsh corgi, Pembroke corgi
			& dachshund, shiba inu
			& dog, \texttt{empty\_string}
			& banana, airplane
			& poodle, beagle, basset hound, chihuahua, french bulldog, pug \\
			
			Pikachu
			& Pikachu character, yellow cartoon mouse
			& Raichu, electric mouse
			& cartoon character, \texttt{empty\_string}
			& chair, ambulance
			& Charmander, Bulbasaur, Squirtle, Kirby, Doraemon, Hello Kitty \\
			
			Hello Kitty
			& white cartoon cat, Hello Kitty character
			& white cat, cartoon cat
			& cartoon character, \texttt{empty\_string}
			& garbage truck, knife
			& My Melody, Kuromi, Doraemon, Pikachu, Mickey Mouse, Snoopy \\
			
			Mickey Mouse
			& Mickey Mouse character, cartoon mouse
			& Minnie Mouse, mouse
			& cartoon character, \texttt{empty\_string}
			& ambulance, knife
			& Donald Duck, Goofy, Pluto, Bugs Bunny, SpongeBob SquarePants, Hello Kitty \\
			
			Eiffel Tower
			& Paris tower, iron tower
			& Tokyo Tower, lattice tower
			& landmark, \texttt{empty\_string}
			& cat, knife
			& Statue of Liberty, Big Ben, Colosseum, Taj Mahal, Golden Gate Bridge, Sydney Opera House \\
			
			garbage truck
			& refuse truck, waste collection truck
			& dump truck, cement mixer
			& vehicle, \texttt{empty\_string}
			& cat, knife
			& ambulance, fire truck, police car, taxi, excavator, front loader \\
			
			ambulance
			& emergency ambulance, medical ambulance
			& white van, delivery van
			& vehicle, \texttt{empty\_string}
			& cat, knife
			& fire truck, police car, taxi, bus, garbage truck, tow truck \\
			
			knife
			& kitchen knife, sharp knife
			& dagger, sword
			& tool, \texttt{empty\_string}
			& cat, ambulance
			& spoon, fork, scissors, hammer, screwdriver, wrench \\
			
			teddy bear
			& stuffed teddy bear, plush bear
			& bear, panda plush
			& toy, \texttt{empty\_string}
			& ambulance, knife
			& stuffed rabbit, plush dog, doll, toy robot, toy car, rubber duck \\
			
			\bottomrule
	\end{tabular}}
\end{table*}

\begin{table*}[t]
	\centering
	\caption{Target suppression and non-target damage under different anchor types. $S_T$ and $R_T$ denote target suppression and target residual, respectively, while $D_X$ denotes non-target damage, where $X\in\{B,R,G\}$ corresponds to the boundary, random, and global concept pools, respectively. Bold and underlined values denote the best and worst results for each metric, respectively.}
	\label{tab:sup_anchor_metrics}
	\small
	\setlength{\tabcolsep}{3.5pt}
	\renewcommand{\arraystretch}{1.15}
	\resizebox{\textwidth}{!}{
		\begin{tabular}{lccccc}
			\toprule
			\textbf{Anchor Type}
			& \shortstack{$S_T$ $\uparrow$}
			& \shortstack{$R_T$ $\downarrow$}
			& \shortstack{$D_B$ $\downarrow$}
			& \shortstack{$D_R$ $\downarrow$}
			& \shortstack{$D_G$ $\downarrow$} \\
			\midrule
			
			Target overlap
			& \underline{0.0423}
			& \underline{0.9577}
			& \textbf{0.0138}
			& 0.0022
			& \textbf{0.0085} \\
			
			Shared visual
			& 0.1008
			& 0.8992
			& 0.0198
			& \textbf{0.0020}
			& 0.0125 \\
			
			Generic / null
			& 0.1292
			& 0.8708
			& 0.0256
			& 0.0028
			& 0.0147 \\
			
			Sibling exclusive
			& 0.1552
			& 0.8448
			& 0.0229
			& 0.0028
			& 0.0127 \\
			
			Far unrelated
			& \textbf{0.2529}
			& \textbf{0.7471}
			& \underline{0.0454}
			& \underline{0.0044}
			& \underline{0.0252} \\
			
			\bottomrule
		\end{tabular}
	}
\end{table*}

\clearpage

\begin{table*}[t]
	\centering
	\caption{Concept configurations for the nine object-erasure targets. For each target, we report the selected SELECT anchor, target synonyms/variants, and the concept retention sets used during model editing. The Boundary Retain Set contains local semantic neighbors preserved by SELECT, whereas the Random Retain Set provides preservation concepts for the fixed-anchor baselines.}
	\label{tabS4}
	\scriptsize
	\renewcommand{\arraystretch}{1.18}
	\setlength{\tabcolsep}{4pt}
	\begin{tabularx}{\textwidth}{
			>{\raggedright\arraybackslash}p{1.7cm}
			>{\raggedright\arraybackslash}p{1.7cm}
			>{\raggedright\arraybackslash}X
			>{\raggedright\arraybackslash}X
			>{\raggedright\arraybackslash}X}
		\toprule
		\textbf{Target}
		&
		\textbf{SELECT Anchor}
		&
		\textbf{Synonyms / Variants}
		&
		\textbf{Boundary Retain Set}
		&
		\textbf{Random Retain Set}
		\\
		\midrule
		
		cat
		&
		dog
		&
		domestic cat; house cat; feline
		&
		raccoon; lion; tiger; bear; wolf
		&
		piano; keyboard; toaster; microphone; projector
		\\
		
		dog
		&
		cat
		&
		domestic dog; canine; pet dog
		&
		wolf; raccoon; lion; tiger; zebra
		&
		Pikachu; garbage truck; banana; statue; backpack
		\\
		
		pig
		&
		sheep
		&
		swine; hog; domestic pig
		&
		cow; goat; donkey; llama; alpaca
		&
		suitcase; toaster; fork; phone; umbrella
		\\
		
		corgi
		&
		dachshund
		&
		Welsh corgi; Pembroke corgi; corgi dog
		&
		poodle; beagle; basset hound; chihuahua; pug
		&
		Pikachu; banana; bicycle; statue; umbrella
		\\
		
		garbage truck
		&
		excavator
		&
		refuse truck; trash truck; waste collection truck
		&
		front loader; grader; ambulance; taxi; fire truck
		&
		cat; pig; Pikachu; guitar; watermelon
		\\
		
		banana
		&
		avocado
		&
		ripe banana; yellow banana; banana fruit
		&
		apple; orange; grape; pineapple; strawberry
		&
		cat; Pikachu; garbage truck; guitar; umbrella
		\\
		
		motorcycle
		&
		scooter
		&
		motorbike; motor cycle; street bike
		&
		bicycle; moped; ATV; car; snowmobile
		&
		cat; pig; Pikachu; banana; backpack
		\\
		
		watermelon
		&
		pumpkin
		&
		watermelon fruit; whole watermelon; watermelon slice
		&
		cantaloupe; honeydew melon; cucumber; pineapple; squash
		&
		scissors; violin; umbrella; bus; train
		\\
		
		Spider-Man
		&
		Batman
		&
		Marvel Spider-Man; Spider-Man superhero
		&
		Iron Man; Superman; Wonder Woman; The Flash; Aquaman
		&
		banana; garbage truck; toothbrush; television; watermelon
		\\
		
		\bottomrule
	\end{tabularx}
\end{table*}

\begin{table*}[t]
	\centering
	\caption{SPEED hyperparameters used for object erasure. The same configuration is used for Null-text, General=\texttt{ground}, and SEC editing.}
	\label{tabS5}
	\small
	\renewcommand{\arraystretch}{1.12}
	\begin{tabular}{lc}
		\toprule
		\textbf{Parameter} & \textbf{Value} \\
		\midrule
		Erasure algorithm & SPEED \\
		aug\_num & 10 \\
		threshold & $1\times10^{-4}$ \\
		retain\_scale & 0.05 \\
		lambda / lamb & 0.0 \\
		disable\_filter & true \\
		chunk\_size & 128 \\
		params & KV \\
		\bottomrule
	\end{tabular}
\end{table*}

\clearpage

\begin{table*}[!t]
	\centering
	
	\begingroup
	\hyphenpenalty=10000
	\exhyphenpenalty=10000
	\emergencystretch=2em
	
	\begin{minipage}[t]{0.485\textwidth}
		\vspace{0pt}
	\begin{tcolorbox}[
		colback=gray!5,
		colframe=black!60,
		boxrule=0.6pt,
		arc=1.5mm,
		title=\textbf{Prompt Template for Object SEC Generation},
		fonttitle=\bfseries,
		fontupper=\small,
		before upper={\raggedright},
		left=2.5mm,
		right=2.5mm,
		top=1.5mm,
		bottom=1.5mm,
		boxsep=0mm
		]
		\label{object_prompt}
		
		Generate \texttt{\{num\_candidates\}} distinct Sibling Exclusive Concepts
		(SECs) for the target object concept
		``\texttt{\{target\_concept\}}''. The alternative concepts must meet
		all these criteria:
		
		\begin{itemize}
			\setlength{\itemsep}{1pt}
			\setlength{\parskip}{0pt}
			\setlength{\parsep}{0pt}
			
			\item Must share the same parent semantic category as
			``\texttt{\{target\_concept\}}'' under the adopted semantic hierarchy,
			and be comparable to the target in semantic granularity.
			
			\item Must remain within the local semantic neighborhood of
			``\texttt{\{target\_concept\}}'', rather than being a distant,
			overly broad, or unrelated concept.
			
			\item Must be mutually exclusive with
			``\texttt{\{target\_concept\}}'' in their core attributes, such that
			the two concepts denote distinct semantic identities rather than
			different names or forms of the same concept.
			
			\item Must not be a synonym, paraphrase, hypernym, hyponym, subclass,
			or direct variant of ``\texttt{\{target\_concept\}}''.
			
			\item Avoid concepts that substantially preserve the defining identity
			of the target and differ only in minor appearance, material, state,
			naming convention, or other non-essential attributes.
			
			\item Each candidate must be unique. Do not generate duplicate or
			near-duplicate concepts.
		\end{itemize}
		
		\textbf{Here is an example:}
		
		Input: ``cat''
		
		Output: ``dog'', ``raccoon'', ``wolf''
		
		Provide \texttt{\{num\_candidates\}} Sibling Exclusive Concepts directly,
		one concept per line.
		
		\textbf{Format requirements:}
		\begin{itemize}
			\setlength{\itemsep}{1pt}
			\setlength{\parskip}{0pt}
			\setlength{\parsep}{0pt}
			
			\item No explanations or additional text.
			\item One concept per line.
			\item Do NOT include ``\texttt{\{target\_concept\}}'' itself,
			its synonyms, variants, subclasses, hypernyms, or hyponyms.
		\end{itemize}
		
		Sibling Exclusive Concepts for
		``\texttt{\{target\_concept\}}'':
		
	\end{tcolorbox}
\end{minipage}
\hfill
\begin{minipage}[t]{0.485\textwidth}
\vspace{0pt}

\begin{tcolorbox}[
	colback=gray!5,
	colframe=black!60,
	boxrule=0.6pt,
	arc=1.5mm,
	title=\textbf{Prompt Template for Celebrity SEC Generation},
	width=\linewidth
	]
	\label{celebrity_prompt}
	\small

	Generate \texttt{\{num\_candidates\}} Sibling Exclusive counterpart
	concepts for the celebrity
	``\texttt{\{target\_concept\}}''. The counterpart
	concepts must meet all these criteria:
	
	\begin{itemize}
		\item Must be specific celebrity identities rather than broad concepts such as
		``person'', ``human'', ``celebrity'', ``a man'', or ``a woman''.
		
		\item Prefer celebrities from the same high-level professional or social-semantic
		domain as ``\texttt{\{target\_concept\}}'' (e.g., politics, music, acting),
		while remaining a distinct identity at the same conceptual granularity.
		
		\item Prefer matching gender presentation when appropriate, while prioritizing
		a semantically comparable but identity-distinct counterpart.
		
		\item Maintain sufficient identity contrast from
		``\texttt{\{target\_concept\}}''. Avoid celebrities that are excessively close
		identity neighbors, such as aliases, relatives, group members, frequent
		collaborators, strongly associated counterparts, or highly confusable identities.
		
		\item The generated names must not be
		``\texttt{\{target\_concept\}}'' or any variation of it, and must not belong
		to the current erasure target set.
		
		\item All generated names must be unique and not repeated.
	\end{itemize}
	
	\textbf{Here is an example:}
	
	Input: ``Leonardo DiCaprio''
	
	Output: ``Tom Hanks'', ``Brad Pitt''.
	
	Provide \texttt{\{num\_candidates\}} celebrity concepts directly,
	one per line.
	
	\textbf{Format requirements:}
	\begin{itemize}
		\item No duplicate names.
		\item No explanations or additional text.
		\item Each name on a new line.
		\item Do NOT include the target concept
		``\texttt{\{target\_concept\}}'' in the list.
	\end{itemize}
	
	Sibling exclusive counterpart celebrities for
	``\texttt{\{target\_concept\}}'':
\end{tcolorbox}
\end{minipage}

\endgroup
\end{table*}

\clearpage

\clearpage

\clearpage

\begin{table*}[t]
	\centering
	\caption{Celebrity erasure configuration. We use 100 identities for erasure and an independent set of 100 identities for retention evaluation.}
	\scriptsize
	\begin{tabular}{p{2cm}p{6cm}p{6cm}}
		\toprule
		\textbf{Concept} & \textbf{Celebrity concepts} & \textbf{Anchor concepts} \\
		\midrule
		Erasure Group & Adam Driver, Adriana Lima, Amber Heard, Amy Adams, Andrew Garfield, Angelina Jolie, Anjelica Huston, Anna Faris,
		Anna Kendrick, Anne Hathaway, Arnold Schwarzenegger, Barack Obama, Beth Behrs, Bill Clinton, Bob Dylan, Bob Marley,
		Bradley Cooper, Bruce Willis, Bryan Cranston, Cameron Diaz, Channing Tatum, Charlie Sheen, Charlize Theron, Chris Evans,
		Chris Hemsworth, Chris Pine, Chuck Norris, Courteney Cox, Demi Lovato, Drake, Drew Barrymore, Dwayne Johnson,
		Ed Sheeran, Elon Musk, Elvis Presley, Emma Stone, Frida Kahlo, George Clooney, Glenn Close, Gwyneth Paltrow,
		Harrison Ford, Hillary Clinton, Hugh Jackman, Idris Elba, Jake Gyllenhaal, James Franco, Jared Leto, Jason Momoa,
		Jennifer Aniston, Jennifer Lawrence, Jennifer Lopez, Jeremy Renner, Jessica Biel, Jessica Chastain, John Oliver, John Wayne,
		Johnny Depp, Julianne Hough, Justin Timberlake, Kate Bosworth, Kate Winslet, Leonardo Dicaprio, Margot Robbie, Mariah Carey,
		Meryl Streep, Mick Jagger, Mila Kunis, Milla Jovovich, Morgan Freeman, Nick Jonas, Nicolas Cage, Nicole Kidman,
		Octavia Spencer, Olivia Wilde, Oprah Winfrey, Paul Mccartney, Paul Walker, Peter Dinklage, Philip Seymour Hoffman, Reese Witherspoon,
		Richard Gere, Ricky Gervais, Rihanna, Robin Williams, Ronald Reagan, Ryan Gosling, Ryan Reynolds, Shia Labeouf,
		Shirley Temple, Spike Lee, Stan Lee, Theresa May, Tom Cruise, Tom Hanks, Tom Hardy, Tom Hiddleston,
		Whoopi Goldberg, Zac Efron, Zayn Malik, Melania Trump
		&
		Lupita-Nyong-o, Tilda-Swinton, Queen-Latifah, Viola-Davis, Lupita-Nyong-o, Keanu-Reeves, Mahershala-Ali, Tilda-Swinton, Hailee-Steinfeld, Lupita-Nyong-o, Danny-DeVito, Greta-Thunberg, Zooey-Deschanel, Lupita-Nyong-o, David-Bowie, Denzel-Washington, Robert-De-Niro, Kurt-Russell, Steve-Carell, Oscar-Isaac, Nelson-Mandela, Lupita-Nyong-o, Danny-DeVito, Lupita-Nyong-o, Tilda-Swinton, James-McAvoy, Lisa-Kudrow, Miley-Cyrus, Kendrick-Lamar, Alicia-Keys, Ryan-Phillippe, Niall-Horan, Sundar-Pichai, Cate-Blanchett, Natalie-Portman, Salvador-Dali, Mahershala-Ali, Lupita-Nyong-o, Keanu-Reeves, Danny-DeVito, Kamala-Harris, Denzel-Washington, Lupita-Nyong-o, Christian-Bale, Tilda-Swinton, Lupita-Nyong-o, Javier-Bardem, Pedro-Pascal, Danny-DeVito, Fergie, Cobie-Smulders, Mark-Ruffalo, Saoirse-Ronan, Rosamund-Pike, Clint-Eastwood, Michael-Fassbender, Selena-Gomez, Pablo-Picasso, Usher, Lupita-Nyong-o, Danny-DeVito, Saoirse-Ronan, Saoirse-Ronan, Melania-Gulić, Frances-McDormand, Albert-Einstein, Lupita-Nyong-o, Christian-Bale, Chiwetel-Ejiofor, Selena-Gomez, Willem-Dafoe, Lupita-Nyong-o, Beyonce, Sofia-Vergara, Samuel-L-Jackson, Sting, John-Cena, Al-Pacino, John-C-Reilly, Keanu-Reeves, Robert-De-Niro, Rowan-Atkinson, Keanu-Reeves, Eddie-Murphy, Denzel-Washington, Jude-Law, Lupita-Nyong-o, Rami-Malek, Raquel-Welch, Quentin-Tarantino, Jim-Shooter, Recep-Tayyip-Erdogan, Brad-Pitt, Gary-Oldman, Lupita-Nyong-o, Benedict-Cumberbatch, Tyler-Perry, Mahershala-Ali, Lupita-Nyong-o, Tilda-Swinton \\
		\midrule
		Retention Group & Aaron Paul, Alec Baldwin, Amanda Seyfried, Amy Poehler, Amy Schumer, Amy Winehouse, Andy Samberg, Aretha Franklin,
		Avril Lavigne, Aziz Ansari, Barry Manilow, Ben Affleck, Ben Stiller, Benicio Del Toro, Bette Midler, Betty White,
		Bill Murray, Bill Nye, Britney Spears, Brittany Snow, Bruce Lee, Burt Reynolds, Charles Manson, Christie Brinkley,
		Christina Hendricks, Clint Eastwood, Countess Vaughn, Dakota Johnson, Dane Dehaan, David Bowie, David Tennant, Denise Richards,
		Doris Day, Dr Dre, Elizabeth Taylor, Emma Roberts, Fred Rogers, Gal Gadot, George Bush, George Takei,
		Gillian Anderson, Gordon Ramsey, Halle Berry, Harry Dean Stanton, Harry Styles, Hayley Atwell, Heath Ledger, Henry Cavill,
		Jackie Chan, Jada Pinkett Smith, James Garner, Jason Statham, Jeff Bridges, Jennifer Connelly, Jensen Ackles, Jim Morrison,
		Jimmy Carter, Joan Rivers, John Lennon, Johnny Cash, Jon Hamm, Judy Garland, Julianne Moore, Justin Bieber,
		Kaley Cuoco, Kate Upton, Keanu Reeves, Kim Jong Un, Kirsten Dunst, Kristen Stewart, Krysten Ritter, Lana Del Rey,
		Leslie Jones, Lily Collins, Lindsay Lohan, Liv Tyler, Lizzy Caplan, Maggie Gyllenhaal, Matt Damon, Matt Smith,
		Matthew Mcconaughey, Maya Angelou, Megan Fox, Mel Gibson, Melanie Griffith, Michael Cera, Michael Ealy, Natalie Portman,
		Neil Degrasse Tyson, Niall Horan, Patrick Stewart, Paul Rudd, Paul Wesley, Pierce Brosnan, Prince, Queen Elizabeth,
		Rachel Dratch, Rachel Mcadams, Reba Mcentire, Robert De Niro & - \\
		\bottomrule
	\end{tabular}
	\label{tabS6}
\end{table*}

\clearpage

\clearpage

\begin{table*}[h]
	\centering
	\caption{The Experiment Setup for Artistic Style Erasure.}
	\scriptsize
	\begin{tabular}{p{1cm}p{7cm}p{7.6cm}}
		\toprule
		\textbf{Group} & \textbf{Artist styles} & \textbf{SELECT anchors} \\
		\midrule
		Erasure Group 
		& 'brent-heighton', 'brett-weston', 'brett-whiteley', 'brian-bolland', 'brian-despain', 'brian-froud', 'brian-k.-vaughan', 'brian-kesinger', 'brian-mashburn', 'brian-oldham', 'brian-stelfreeze', 'brian-sum', 'briana-mora', 'brice-marden', 'bridget-bate-tichenor', 'briton-rivière', 'brooke-didonato', 'brooke-shaden', 'brothers-grimm', 'brothers-hildebrandt', 'bruce-munro', 'bruce-nauman', 'bruce-pennington', 'bruce-timm', 'bruno-catalano', 'bruno-munari', 'bruno-walpoth', 'bryan-hitch', 'butcher-billy', 'c.-r.-w.-nevinson', 'cagnaccio-di-san-pietro', 'camille-corot', 'camille-pissarro', 'camille-walala', 'canaletto', 'candido-portinari', 'carel-willink', 'carl-barks', 'carl-gustav-carus', 'carl-holsoe', 'carl-larsson', 'carl-spitzweg', 'carlo-crivelli', 'carlos-schwabe', 'carmen-saldana', 'carne-griffiths', 'casey-weldon', 'caspar-david-friedrich', 'cassius-marcellus-coolidge', 'catrin-welz-stein', 'cedric-peyravernay', 'chad-knight', 'chantal-joffe', 'charles-addams', 'charles-angrand', 'charles-blackman', 'charles-camoin', 'charles-dana-gibson', 'charles-e.-burchfield', 'charles-gwathmey', 'charles-le-brun', 'charles-liu', 'charles-schridde', 'charles-schulz', 'charles-spencelayh', 'charles-vess', 'charles-francois-daubigny', 'charlie-bowater', 'charline-von-heyl', 'chaïm-soutine', 'chen-zhen', 'chesley-bonestell', 'chiharu-shiota', 'ching-yeh', 'chip-zdarsky', 'chris-claremont', 'chris-cunningham', 'chris-foss', 'chris-leib', 'chris-moore', 'chris-ofili', 'chris-saunders', 'chris-turnham', 'chris-uminga', 'chris-van-allsburg', 'chris-ware', 'christian-dimitrov', 'christian-grajewski', 'christophe-vacher', 'christopher-balaskas', 'christopher-jin-baron', 'chuck-close', 'cicely-mary-barker', 'cindy-sherman', 'clara-miller-burd', 'clara-peeters', 'clarence-holbrook-carter', 'claude-cahun', 'claude-monet', 'clemens-ascher'
		& 'Abstract-Expressionism-style', 'Academic-Art-style', 'structured fauvist landscape painting style', 'Aestheticism-style', 'streamlined art-deco robot illustration style', 'American-Realism-style', 'Art-Brut-style', 'Art-Deco-style', 'Art-Nouveau-style', 'documentary fine-art portrait photography style', 'Ashcan-School-style', 'hard-surface science-fiction concept art style', 'Bauhaus-style', 'Biopunk-style', 'Byzantine-Art-style', 'Celtic-Art-style', 'Chiaroscuro-style', 'Color-Field-Painting-style', 'Conceptual-Art-style', 'Constructivism-style', 'Cubism-style', 'Cyberpunk-Art-style', 'Dadaism-style', 'De-Stijl-style', 'Deconstructivism-style', 'Digital-Art-style', 'Dutch-Golden-Age-Painting-style', 'Earth-Art-style', 'Expressionism-style', 'Fauvism-style', 'metaphysical realist figurative painting style', 'Folk-Art-style', 'Minimalist-vector-illustration-style', 'Tachisme-style', 'Glitch-Art-style', 'Gothic-Art-style', 'Graffiti-Art-style', 'Hard-Edge-Painting-style', 'Harlem-Renaissance-Art-style', 'High-Renaissance-style', 'Hudson-River-School-style', 'Hyperrealism-style', 'Impressionism-style', 'Installation-Art-style', 'Islamic-Architecture-style', 'Japonisme-style', 'Kinetic-Art-style', 'De-Stijl-style', 'Constructivism-style', 'Light-and-Space-movement-style', 'Bauhaus-style', 'Luminism-style', 'Lyrical-Abstraction-style', 'Magic-Realism-style', 'Cubism-style', 'Maximalism-style', 'Medieval-Art-style', 'Memphis-Design-style', 'Metaphysical-Art-style', 'Fauvism-style', 'Modernism-style', 'Mughal-Painting-style', 'Dadaism-style', 'Neoclassicism-style', 'Neo-Dada-style', 'Neo-Expressionism-style', 'Neo-Geo-style', 'Neo-Impressionism-style', 'Suprematism-style', 'Op-Art-style', 'Orphism-style', 'Outsider-Art-style', 'Performance-Art-style', 'De-Stijl-style', 'Photorealism-style', 'Pixel-Art-style', 'Pointillism-style', 'Pop-Art-style', 'Post-Impressionism-style', 'Postmodernism-style', 'Precisionism-style', 'Pre-Raphaelite-Brotherhood-style', 'flat-color mid-century architectural illustration style', 'Realism-style', 'Regionalism-style', 'Rococo-style', 'cinematic fantasy landscape concept art style', 'atmospheric science-fiction environment concept art style', 'Russian-Futurism-style', 'Social-Realism-style', 'Color-Field-Painting-style', 'Street-Art-style', 'Suprematism-style', 'Surrealism-style', 'decorative stained-glass illustration style', 'Tachisme-style', 'Glitch-Art-style', 'Ukiyo-e-style', 'Vaporwave-Art-style', 'Vorticism-style'
		\\
		\midrule
		Retention Group & 'A.J.Casson', 'Aaron Douglas', 'Aaron Horkey', 'Aaron Jasinski', 'Aaron Siskind', 'Abbott Fuller Graves', 'Abbott Handerson Thayer', 'Abdel Hadi Al Gazzar', 'Abed Abdi', 'Abigail Larson', 'Abraham Mintchine', 'Abraham Pether', 'Abram Efimovich Arkhipov', 'Adam Elsheimer', 'Adam Hughes', 'Adam Martinakis', 'Adam Paquette', 'Adi Granov', 'Adolf Hiremy-Hirschl', 'Adolph Gottlieb', 'Adolph Menzel', 'Adonna Khare', 'Adriaen van Ostade', 'Adriaen van Outrecht', 'Adrian Donoghue', 'Adrian Ghenie', 'Adrian Paul Allinson', 'Adrian Smith', 'Adrian Tomine', 'Adrianus Eversen', 'Afarin Sajedi', 'Affandi', 'Aggi Erguna', 'Agnes Cecile', 'Agnes Lawrence Pelton', 'Agnes Martin', 'Agostino Arrivabene', 'Agostino Tassi', 'Ai Weiwei', 'Ai Yazawa', 'Akihiko Yoshida', 'Akira Toriyama', 'Akos Major', 'Akseli Gallen-Kallela', 'Al Capp', 'Al Feldstein', 'Al Williamson', 'Alain Laboile', 'Alan Bean', 'Alan Davis', 'Alan Kenny', 'Alan Lee', 'Alan Moore', 'Alan Parry', 'Alan Schaller', 'Alasdair McLellan', 'Alastair Magnaldo', 'Alayna Lemmer', 'Albert Benois', 'Albert Bierstadt', 'Albert Bloch', 'Albert Dubois-Pillet', 'Albert Eckhout', 'Albert Edelfelt', 'Albert Gleizes', 'Albert Goodwin', 'Albert Joseph Moore', 'Albert Koetsier', 'Albert Kotin', 'Albert Lynch', 'Albert Marquet', 'Albert Pinkham Ryder', 'Albert Robida', 'Albert Servaes', 'Albert Tucker', 'Albert Watson', 'Alberto Biasi', 'Alberto Burri', 'Alberto Giacometti', 'Alberto Magnelli', 'Alberto Seveso', 'Alberto Sughi', 'Alberto Vargas', 'Albrecht Anker', 'Albrecht Durer', 'Alec Soth', 'Alejandro Burdisio', 'Alejandro Jodorowsky', 'Aleksey Savrasov', 'Aleksi Briclot', 'Alena Aenami', 'Alessandro Allori', 'Alessandro Barbucci', 'Alessandro Gottardo', 'Alessio Albi', 'Alex Alemany', 'Alex Andreev', 'Alex Colville', 'Alex Figini', 'Alex Garant' & - \\
		\bottomrule
	\end{tabular}
	\label{tabS7}
\end{table*}

\clearpage

\begingroup
\hyphenpenalty=10000
\exhyphenpenalty=10000
\emergencystretch=2em

\begin{tcolorbox}[
	colback=gray!5,
	colframe=black!60,
	boxrule=0.6pt,
	arc=1.5mm,
	title=\textbf{Prompt Template for Artist-Style SEC Generation},
	fonttitle=\bfseries,
	fontupper=\small,
	before upper={\raggedright},
	left=2.5mm,
	right=2.5mm,
	top=1.5mm,
	bottom=1.5mm,
	boxsep=0mm
	]
	\label{style_prompt}
	
	Generate \texttt{\{num\_candidates\}} distinct Sibling Exclusive Counterpart
	artistic styles for the target artistic style
	``\texttt{\{target\_concept\}}''. The alternative artistic styles must meet
	all these criteria:
	
	\begin{itemize}
		\setlength{\itemsep}{1pt}
		\setlength{\parskip}{0pt}
		\setlength{\parsep}{0pt}
		
		\item Must belong to the same high-level semantic domain of artistic styles
		and be comparable to ``\texttt{\{target\_concept\}}'' in stylistic granularity,
		while remaining visually and stylistically distinct.
		
		\item Must represent artistic styles that are visually and stylistically
		different from ``\texttt{\{target\_concept\}}''.
		
		\item The generated artistic styles must not be
		``\texttt{\{target\_concept\}}'' or any variation of it.
		
		\item Focus on maintaining sufficient visual contrast and stylistic differences
		from the target concept while remaining within the same high-level
		artistic-style domain. Avoid styles that merely paraphrase the target style
		or preserve its dominant stylistic signature, such as closely related
		variants sharing the same characteristic brushwork, palette, composition,
		texture, medium, or motifs.
	\end{itemize}
	
	\textbf{Here is an example:}
	
	Input: ``Van Gogh's style''
	
	Output: ``Photorealism-style'', ``De-Stijl-style''.
	
	Provide \texttt{\{num\_candidates\}} artistic styles directly,
	one per line.
	
	\textbf{Format requirements:}
	\begin{itemize}
		\setlength{\itemsep}{1pt}
		\setlength{\parskip}{0pt}
		\setlength{\parsep}{0pt}
		
		\item No explanations or additional text.
		\item Each style/artist on a new line.
		\item Do NOT include the target artistic style
		``\texttt{\{target\_concept\}}'' in the list.
	\end{itemize}
	
	Sibling exclusive counterpart artistic styles for
	``\texttt{\{target\_concept\}}'':
	
\end{tcolorbox}

\endgroup

\begin{table*}[htbp]
	\centering
	\caption{Sibling-Exclusive Concepts (SECs)}
	\label{tabS8}
	\begin{tabular}{l|p{7cm}}
		\toprule
		Concept & Sibling-Exclusive Concepts (SECs) \\
		\midrule
		Nudity/naked & "a firefighter in full gear", \\
		& "an arctic explorer in a heavy parka", \\
		& "a beekeeper in a protective suit", \\
		& "a welder in a leather apron and helmet", \\
		& "a scientist in a cleanroom bunny suit", \\
		& "a scuba diver in a full wetsuit", \\
		& "a construction worker in a high-visibility jacket and pants", \\
		& "a person in a thick, woolen winter coat", \\
		& "a motorcyclist in full leather racing suit", \\
		& "a gardener in overalls and long sleeves" \\
		\midrule
		Sexual/erotic & "a professional architect reviewing blueprints", \\
		& "a dentist examining a patient's teeth", \\
		& "a pharmacist filling a prescription", \\
		& "a museum curator arranging an exhibit", \\
		& "a software engineer attending a team meeting", \\
		& "a geologist examining a rock formation", \\
		& "a pilot in a cockpit preparing for takeoff", \\
		& "a historian giving a lecture in a university", \\
		& "a tailor measuring fabric in a workshop", \\
		& "a gardener planting flowers in a public park" \\
		\bottomrule
	\end{tabular}
	\label{tab12}
\end{table*}

\clearpage

\end{document}